\documentclass[12pt]{article}
\usepackage[utf8]{inputenc}
\usepackage[T1]{fontenc}
\usepackage{physics}
\usepackage{algorithm2e}

\usepackage{graphicx}
\usepackage{authblk}
\usepackage{amsmath, amsfonts}
\usepackage{url}

\usepackage[letterpaper, left=2cm,
right=2cm, top=2.5cm,bottom=2.5cm,dvips]{geometry}

\usepackage{float}

\DeclareMathOperator{\rk}{rank}

\usepackage[font=footnotesize, labelfont=bf, justification=justified]{caption}
\usepackage{subcaption}
\title{\textbf{Robust estimation of the intrinsic dimension of data sets with quantum cognition machine learning}}

\author[1,2,*]{Luca Candelori}
\author[3,+]{Alexander G. Abanov}
\author[1,+]{Jeffrey Berger}
\author[4,+]{Cameron J. Hogan}
\author[1,+]{Vahagn Kirakosyan}
\author[1,+]{Kharen Musaelian}
\author[1,+]{Ryan Samson}
\author[1,+]{James E. T. Smith}
\author[1,+]{Dario Villani}
\author[4,+]{Martin T. Wells}
\author[5,6,+]{Mengjia Xu}

\affil[1]{Qognitive, Inc., Miami Beach, FL 33139, USA}
\affil[2]{Wayne State University, Department of Mathematics, Detroit, MI 48202}
\affil[3]{Stony Brook University, Department of Physics and Astronomy, Stony Brook, NY 11790}
\affil[4]{Cornell University, Department of Statistics and Data Science, Ithaca, NY 14853}
\affil[5]{New Jersey Institute of Technology, Department of Data Science, Newark, NJ 07102}
\affil[6]{Massachusetts Institute of Technology, Center for Brains, Minds and Machines, Cambridge, MA 02139}

\affil[*]{luca.candelori@qognitive.io, candelori@wayne.edu}
\affil[+]{these authors contributed equally to this work and are listed alphabetically}

\date{}

\begin{document}

\maketitle

\begin{abstract}
We propose a new data representation method based on Quantum Cognition Machine Learning and apply it to manifold learning, specifically to the estimation of intrinsic dimension of data sets. The idea is to learn a representation of each data point as a quantum state, encoding both local properties of the point as well as its relation with the entire data. Inspired by ideas from quantum geometry, we then construct from the quantum states a point cloud equipped with a quantum metric. The metric exhibits a spectral gap whose location corresponds to the intrinsic dimension of the data. The proposed estimator is based on the detection of this spectral gap. When tested on synthetic manifold benchmarks, our estimates are shown to be robust with respect to the introduction of point-wise Gaussian noise. This is in contrast to current state-of-the-art estimators, which tend to attribute artificial ``shadow dimensions'' to noise artifacts, leading to overestimates. This is a significant advantage when dealing with real data sets, which are inevitably affected by unknown levels of noise. We show the applicability and robustness of our method on real data, by testing it on the ISOMAP face database, MNIST, and the Wisconsin Breast Cancer Dataset.  
\end{abstract}

\flushbottom
\maketitle

\thispagestyle{empty}

\section*{Introduction}

When data is characterized by a large number of features (e.g., zip code, annual income, age, credit card spend, etc. for borrowers; cholesterol, blood pressure, BMI, blood pressure, etc. for patients; or the latent and dependent variables), it tends to lie on a surface that has a smaller dimensionality than the full feature space \cite{bishop1995neural}. Finding this low-dimensional surface is often referred to as {\em manifold learning}. The smaller dimensionality reflects the underlying latent structures in the data, correlations, and a variety of nonlinear relationships \cite{johnson1984extensions, donoho2000high}. Furthermore, data points whose feature vectors are close together should possess similar properties. For example, in a supervised regression problem the output/target variables are expected to depend smoothly on the input variables. These characteristics of real data suggest that any given dataset consisting of $D$ features lies entirely on a smooth manifold $M\subseteq \mathbb{R}^D$ of dimension $d \ll D$, the {\em intrinsic dimension} of the data \cite{bishop1995neural}. This number represents the minimal number of parameters required to characterize the data. Knowledge of the intrinsic dimension $d$ can be used to effectively choose a target space for dimension-reduction models (such as PCA, Isomap, t-SNE, etc.)  or to compress deep neural networks while maintaining the performance \cite{id-DNN}. Intrinsic dimension estimation is also widely used in network analysis \cite{macocco2024intrinsic,xu2020new}, complex materials \cite{zhou2021local} and health sciences \cite{varghese2023global}.

One of the main challenges for manifold learning is the inevitable presence of noise in real data.  A typical ``global'' approach is to impose a functional form (e.g. PCA where the manifold is assumed to be linear) and to assume that the error between the manifold approximation and the actual data is noise, which is then analyzed. Imposing a functional relation immediately gives estimates for the intrinsic dimension, which tend to be robust to the introduction of additional noise. However, when the data manifold $M$ has a lot of curvature, linear methods will fail. The problem can be somewhat alleviated by sampling locally around each data point, assuming that at a sufficiently small scale all manifolds are close to being linear \cite{lPCA, cangelosi, Little}. Indeed, all current state-of-the-art  intrinsic dimension estimators (some of which we describe below) are ``local'', producing estimates that are based on a local sampling around each data point. Such techniques are designed and benchmarked against highly dimensional, highly curved manifolds. While they perform reasonably well in this ideal setup, they often tend to fall apart when noise is re-introduced into the data. Indeed, local methods cannot distinguish {\em shadow} dimensions that are transversal to the data manifold, and that are only artifacts created by the noise, leading to significant overestimates of intrinsic dimension. 

In this paper, we propose a new data representation and manifold learning technique based on Quantum Cognition Machine Learning (QCML) \cite{QCML} and quantum geometry \cite{Ishiki, Schneiderbauer_2016, Steinacker_2021}. The main idea is to create a (non-commutative) quantum model for the data manifold itself, from which we can estimate important geometric features, such as intrinsic dimension. Picking a quantum model is similar to what is done in linear methods, in the sense that a functional relation is imposed on the data. But in contrast to linear methods we {\em learn} the model from the data, and we make no assumptions about the underlying distribution. Our method gives local estimates of intrinsic dimension at every data point, but it also takes into account the global geometry of the data manifold $M$. To this end, we are able to develop a manifold approximation method that is both robust to noise and flexible enough to capture non-linear geometric features of the data manifold. 

In addition to the aforementioned local PCA methods, current state-of-the-art intrinsic dimension estimators measure statistics related to the density of neighbors lying within a certain radius $r$ from a data point $x$, and express these statistics as functions of intrinsic dimension (CorrInt \cite{Grassberger-Procaccia}, MLE \cite{Levina2004MaximumLE}, DANCo \cite{CERUTI-DANCo}, TwoNN \cite{Facco-2NN}). These methods do not make any linearity assumption about the data, but do require the data to be dense in small patches around any given point. As is well-known, this requirement is fundamentally incompatible with the {\em curse of dimensionality}\cite{curse_of_dim}, usually occurring in dimensions $d \gg 10$, and indeed these methods tend to underestimate the intrinsic dimension $d$ when $d$ is large. The overestimation effect induced by noise combined with the underestimation effect induced by the curse of dimensionality often results in completely unreliable intrinsic dimension estimates. 

Compared to existing intrinsic dimension estimators, which are all based on local sampling of the data, our method first learns a model for the entire data manifold $M$, as a semi-classical limit of a quantum matrix configuration (in the sense of quantum geometry \cite{Ishiki, Schneiderbauer_2016,Steinacker_2021}). In particular, given a data set $X$ containing $D$ features, we propose to train $D$ quantum observables $A = \{A_1, \ldots, A_D\}$ (i.e. a {\em matrix configuration}) as it is done in QCML \cite{QCML}. We then calculate from $A$ a {\em point cloud} approximation $X_A$ to the actual data manifold $M$. Each element $x\in X_A$ of the point cloud represents the expected position in feature space of its corresponding data point, and it comes with a ``cloud'' of uncertainty around its actual position whose shape is determined by the quantum fluctuations of the matrix configuration. The point $x$ is further equipped with a {\em quantum metric} $g(x)$, which is a $D\times D$ real symmetric positive semi-definite matrix. This metric, already considered by physicists \cite{Ishiki, Schneiderbauer_2016}, encodes much of the local geometry of the data manifold; it can be shown that its rank in particular is approximately equal to the intrinsic dimension of $M$, and that its non-zero eigenvalues are all close to 1. Therefore, intrinsic dimension estimates can be given by detecting the spectral gap of the quantum metric, separating the zero eigenvalues from the non-zero eigenvalues that are close to 1. 

We test our intrinsic dimension estimator on both synthetic and real data sets, following the benchmarking framework proposed in ref. \cite{Campadelli} and implemented in the \texttt{scikit-dimension} Python package \cite{scikit-dimension}. In addition to this standard framework, we stress-test our estimates by introducing increasing levels of Gaussian noise into the data, and compare the results with other state-of-the-art techniques.  In all of our testing, higher levels of noise increasingly degrade the quality of the point cloud approximation $X_A$, and the spectral gap detection in the quantum metric becomes increasingly difficult. However, they do not qualitatively alter the intrinsic dimension estimation. This stands in marked contrast to other intrinsic dimension estimators that we tested, whose estimates are highly sensitive to even small amounts of noise.  

\section*{Results}

\subsection*{Quantum geometry in data analysis}
Consider a $t \times D$ data set $X$ containing $t$ data points $x_1, \ldots, x_t$, where each data point $x_i$ consists of $D$-dimensional vector of data features $x_i = (a^1_i, \ldots, a^D_i)$. We assume that $X$ lies entirely on a smooth manifold $M$, called the {\em data manifold}, of intrinsic dimension $d < D$. We further assume that the $D$ features of the data extend to smooth functions in $C^{\infty}(M)$, giving the coordinates of an embedding $a^k:M\hookrightarrow \mathbb{R}^D$ of the data manifold into $D$-dimensional Euclidean space. In quantum geometry, the commutative algebra $C^{\infty}(M)$ of smooth functions on a manifold is replaced by the non-commutative algebra of Hermitian operators on a $N$-dimensional Hilbert space \cite{Steinacker_book, oxidation}. For the purposes of this work, any set $A = \{ A_1, \ldots, A_D\}$ consisting of $D$ Hermitian $N\times N$ matrices is called a {\em matrix configuration}, and can be viewed as a non-commutative avatar of the $D$ coordinate functions $a^k$ on a manifold $M\hookrightarrow \mathbb{R}^D$. Typically in physics, the matrix configuration $A$ is given by a quantum theory and the goal is to construct a symplectic manifold $M\hookrightarrow \mathbb{R}^D$, so that $A$ represents a quantization of the coordinate functions $x^k$ giving the embedding; that is, a compatibility between the Poisson bracket on $M$ and the commutator bracket on $A$ is required, among other conditions. 

In the context of data analysis, the situation is reversed: $M$ is given by the data manifold, and we propose instead to learn a suitable matrix configuration $A$, reflecting as much of the geometry of $M$ as possible. We do so through the formalism of {\em quasi-coherent states} \cite{Ishiki, Schneiderbauer_2016}. Recall that in quantum mechanics a {\em state} is a vector of unit norm in a Hilbert space, and is represented in bra-ket notation by a ket $\ket{\psi}$. The inner product of two states $\ket{\psi_1}, \ket{\psi_2}$ is represented by a bra-ket $\braket{\psi_1}{\psi_2}$. The {\em expectation value} of a Hermitian operator $A$ on a state $\ket{\psi}$ is denoted by $\expval{A}{\psi} = \braket{A\psi}{\psi} = \braket{\psi}{A\psi}$, representing the expected outcome of the measurement corresponding to $A$ on the state $\ket{\psi}$. For any state $\ket{\psi}$ in $N$-dimensional Hilbert space and an $N\times N$ matrix configuration $A = \{A_1, \ldots, A_D\}$, define the state's {\em position} vector by 
\[
A(\psi) = \left( \expval{A_1}{\psi}, \ldots, \expval{A_D}{\psi}   \right) \in \mathbb{R}^D
\]
and the state's {\em variance} (or {\em quantum fluctuation}) $\sigma^2(\psi)$  by 
\[
\sigma_k^2(\psi) = \expval{A^2_k}{\psi}  - \expval{A_k}{\psi}^2, \quad 
\sigma^2(\psi) = \sum_{k=1}^D \sigma_k^2(\psi) \in \mathbb{R}.
\]
Intuitively, the matrix configuration $A$ assigns to each quantum state $\ket{\psi}$ a point $A(\psi)$ in Euclidean space $\mathbb{R}^D$, together with a ``cloud'' around it representing the uncertainty of the measurement of the point's position in space. In this picture, $A(\psi)$ represents the center of the cloud, while $\sigma(\psi)$ is a statistical measure of the cloud's size. 

Now for any data point $x = (a_k)\in \mathbb{R}^D$, we want to construct a quantum state $\psi_0(x)$ reflecting not only the absolute position of $x$ within feature space, but also its relation to all the other points in the data set $X$. To do so, consider the {\em error Hamiltonian}
\begin{equation}
\label{eqn:errorHamiltonian}
H(x) = \frac{1}{2} \sum_{k=1}^D (A_k - a_k\cdot I_N)^2,
\end{equation}
a positive semi-definite Hermitian operator. We will assume throughout the article that all the eigenvalues of $H(x)$ are distinct, so that all the eigenspaces are one-dimensional. Practically, when dealing with real numerical data degeneracies of $H(x)$ do not play any role. Denote by $E_0(x), \ldots, E_n(x)$ the eigenvalues of $H(x)$, listed in increasing order, and let $\ket{\psi_0(x)}, \ldots, \ket{\psi_n(x)}$ be corresponding choices of normalized eigenvectors, or {\em eigenstates}. By assumption, all eigenstates are uniquely defined up to multiplication by a phase factor  $e^{i\theta}, \theta \in \mathbb{R}$. For each $x$, an eigenstate $\ket{\psi_0(x)}$ associated to the lowest eigenvalue of $H(x)$ is called a {\em quasi-coherent state} of $x$. A simple calculation shows that 
\begin{equation}
\label{eqn:ground}
E_0(x) = \frac{1}{2}\lVert A(\psi_0(x)) - x\rVert^2  + \frac{1}{2}\sigma^2(\psi_0(x)),
\end{equation}
so that the lowest eigenvalue (i.e. the {\em ground state energy}) of the error Hamiltonian can be broken down into two contributions: the squared distance between $x$ and the position of its corresponding quasi-coherent state, and the quantum fluctuation of the quasi-coherent state itself. This is analogous to the bias-variance breakdown of the mean-squared error loss function. We can now train a matrix configuration $A$ so as to minimize the combined loss function \eqref{eqn:ground} for all data points $x\in X$. In this way, the matrix configuration captures global features of the data, which are then reflected into the ground state $\psi_0(x)$, for each $x\in X$.

% Given now a data manifold $M \hookrightarrow \mathbb{R}^D$ and a $N\times N$ matrix configuration $A = \{A_1, \ldots, A_D\}$ let 
% \begin{equation}
% \label{eqn:fuzzy_manifold}
% M_A = \{ A(\psi_0(x)) : x \in M\}
% \end{equation}
% be the {\em cloud manifold} approximation of $M$. In this approximation each point of $M$ has been replaced by the center of its corresponding point cloud, so that $M_A$ comes equipped with a degree of uncertainty, or `fuzzy-ness', surrounding every point. 

From the trained matrix configuration $A$, we may then calculate the {\em point cloud}
\begin{equation}
\label{eqn:point_cloud}
X_A = \{ A(\psi_0(x)) : x \in X\} \subseteq \mathbb{R}^D,
\end{equation}
which can be viewed as an approximate sampling of the data manifold $M$. The original data points $x \in X$ may contain noise, missing features, or otherwise deviate substantially from the idealized underlying data manifold $M$. By choosing an appropriate matrix configuration $A$, capturing enough global information about the data, the set $X_A$ turns out to be much closer to $M$ than the original data set $X$. Key geometric features of the data manifold, such as the intrinsic dimension $d$, can be recovered from $X_A$ in a way that is robust to noise and other artifacts.

\subsection*{Quantum Cognition Machine Learning}

Training a matrix configuration $A$ on a data set $X$ is the optimization problem forming the basis of Quantum Cognition Machine Learning (QCML) \cite{QCML}. QCML has been developed independently of quantum geometry, and this is the first work pointing out the relation between the two. In the original formulation of QCML, a matrix configuration $A$ is trained so as to minimize the aggregate energy loss function \eqref{eqn:ground} across all data. In the present context, minimizing energy sometimes has the undesired effect of training $A$ so that the aggregate quantum fluctuation $\sum_{x\in X} \sigma^2(\psi(x))$ goes to zero, forcing all the matrices $A_1, \ldots, A_D$ in the matrix configuration to commute. A commutative matrix configuration is highly undesirable. It produces a point cloud approximation $X_A$ consisting of $N$ points, corresponding to the positions of the $N$ common eigenstates of the matrix configuration, with no point cloud around them. Indeed, it can be shown that $X_A$ in this case consists of a $N$-means clustering of the data set $X$, and is therefore entirely classical\cite{poggio_kmeans}.

Instead, in this work we train the matrix configuration $A = \{A_1, \ldots, A_D\}$ on the data set $X$ by minimizing the mean squared distance between the data set $X$ and the point cloud $X_A$, i.e. by finding 
\begin{equation}
\label{eqn:matrix_config_optimization}
A = \displaystyle\mathrm{argmin}_{B=\{B_1, \ldots, B_D\}} \left( \sum_{x\in X} {\lVert B(\psi_0(x)) - x\rVert^2}\right),
\end{equation}
where the minimum is taken over the space of all $D$-tuples of $N\times N$ Hermitian matrices. The optimization \eqref{eqn:matrix_config_optimization} can be tackled efficiently using gradient descent methods, similar to those employed in the state-of-the-art machine learning models. In our study, we find $A$ by implementing the optimization problem as a custom layer in PyTorch~\cite{PyTorch}.

Note that the choice of loss function in \eqref{eqn:matrix_config_optimization} corresponds to the ``squared-bias'' term in the bias-variance decomposition of the energy functional $E_0(x)$ in \eqref{eqn:ground}. We do not minimize the quantum fluctuation, or ``variance'' term. Indeed, while the bias term is in general unbounded, the quantum fluctuation $\sigma^2(x)$ has a simple bound in terms of the matrix configuration $A$ only (i.e. independent of $x$), given by 
\[
\sigma^2(x) \leq \sum_{k=1}^D (\mu_k - m_k)^2 \leq \frac{D}{4}(\mu- m)^2,
\]
where $\mu_k$ (resp. $m_k$) is the highest (resp. lowest) eigenvalue of $A_k$ and $\mu = \max_k \mu_k$ (resp. $m = \min_k m_k)$. This bound has an elementary proof similar to Popoviciu's inequality \cite{popoviciu1935equations} on variances. Note that the eigenvalues of $A_k$ correspond to possible measurement outcomes of the $k$-th coordinate of the position of a point $x$. Therefore, if we train $A$ so that the positions $X_A$ are close to a compact data set $X$, we expect the quantum fluctuation to be commensurate with the average noise level in the data $X$. This is indeed what we observe in practice.

It is also possible to modify the loss function in  \eqref{eqn:matrix_config_optimization} by adding back the quantum fluctuation term with a weight $w \in \mathbb{R}_{\geq 0}$, a tunable hyperparameter, 
\begin{equation}
\label{eqn:matrix_config_optimization_with_variance}
A = \displaystyle\mathrm{argmin}_{B=\{B_1, \ldots, B_D\}} \left( \sum_{x\in X} {\lVert B(\psi_0(x)) - x\rVert^2} + w\cdot \sigma^2(x) \right).
\end{equation}
In this way, the choice $w=0$ recovers the bias-only loss function  \eqref{eqn:matrix_config_optimization} while $w=1$ corresponds to the original energy loss \eqref{eqn:ground}. In applications, small non-zero values of $w$ may lead to more robust point cloud approximations $X_A$, especially in the presence of significant amounts of noise. 

It is also possible in principle to replace the error Hamiltonian \eqref{eqn:errorHamiltonian} with the Dirac operator defined in ref. \cite{Schneiderbauer_2016}. The advantage of using the Dirac operator is that the energy loss is allowed to reach zero without the matrix configuration $A$ being necessarily commutative. Equivalently, the quasi-coherent states in this case are zero modes. However, the Hilbert space dimension required by the Dirac operator scales  exponentially in the number of features $D$, and this is not practical when dealing with data sets containing a large number of features. 

\subsection*{Intrinsic dimension estimation}

Suppose now that a matrix configuration $A$ has been trained from a data set $X$ as in \eqref{eqn:matrix_config_optimization}, so that the data manifold $M$, by construction, lies within a region of $\mathbb{R}^D$ where the energy functional $E_0(x)$ is near-minimal and it has minimal variation (assuming that the quantum fluctuation term in \eqref{eqn:ground} is not too large). We may then apply the technique described in ref. \cite{Schneiderbauer_2016} to calculate the intrinsic dimension of $M$. In particular, from formula \eqref{eqn:ground}, we see that as $x$ moves away from the manifold $M$ then the energy $E_0(x)$ increases like the squared distance from $x$ to $M$, while in the directions tangent to $M$ the energy is approximately constant. This means that the Hessian matrix of the energy functional at $x$ should exhibit a clear spectral gap between the lowest $d = \dim M$ eigenvalues, corresponding to the directions tangent to $M$ and near zero, and the highest $D-d$ eigenvalues, of order one and corresponding to the directions that point away from $M$. Detecting the exact location of the spectral gap is therefore equivalent to estimating the intrinsic dimension of $M$. 

This observation can be turned into an algorithm for estimating intrinsic dimension. First, the Hessian matrix of the energy functional can be computed in terms of the matrix configuration $A$, using perturbation theory. Its entries are given by the formula 

\begin{equation}
\label{eqn:Hessian}
\pdv{E_0}{x_{\mu}}{x_{\nu}} = \delta_{\mu\nu} - 2\sum_{n=1}^{N-1} \Re{\frac{\mel{\psi_0(x)}{A_{\mu}}{\psi_n(x)}\mel{\psi_n(x)}{A_{\nu}}{\psi_0(x)}}{E_n(x) - E_0(x)}}, \quad \mu,\nu = 1, \ldots, D
\end{equation}
where, as before, we write $\psi_n(x)$ and $E_n(x)$ for the eigenstates and energies of the error Hamiltonian $H(x)$ given by \eqref{eqn:errorHamiltonian}. Notice that \eqref{eqn:Hessian} is exact, despite being derived using perturbation theory. In detecting the spectral gap, it is more convenient to consider the second term of \eqref{eqn:Hessian} only, a real symmetric $D\times D$ matrix $g(x)$ whose entries are given by 

\begin{equation}
\label{eqn:quantum_metric}
g_{\mu\nu}(x) = 2\sum_{n=1}^{N-1} \Re{\frac{\mel{\psi_0(x)}{A_{\mu}}{\psi_n(x)}\mel{\psi_n(x)}{A_{\nu}}{\psi_0(x)}}{E_n(x) - E_0(x)}}, \quad \mu,\nu = 1, \ldots, D. 
\end{equation}

It can be easily shown that the matrix $g(x)$ is positive semi-definite, and in the context of matrix geometry it is called the {\em quantum metric}\cite{Schneiderbauer_2016,Steinacker_2021,oxidation}. Indeed, it can be viewed as an approximate Riemannian metric on the data manifold $M$, and  
$\dim_x M \approx \rk g(x)$. We could in principle apply this formula to estimate intrinsic dimension, by training a matrix configuration $A$ on the data set $X$ and then estimate the ranks of $g(x)$, without making use of the point cloud $X_A$ defined in \eqref{eqn:point_cloud}. However, as noted in ref. \cite{Schneiderbauer_2016}, much clearer spectral gaps emerge in practice when calculating the quantum metric on $X_A$. This is because $X_A$, as noted earlier, is much more robust to noise and to small perturbations of the data manifold. Note that since the matrix configuration $A$ was trained in such a way as to minimize the squared distance between $X$ and $X_A$, it is reasonable to assume that the intrinsic dimensions of both data sets are equal. 

The algorithm for estimating intrinsic dimension can be summarized as follows. 

\RestyleAlgo{ruled}
\LinesNumbered
\begin{algorithm}[H]%[!ht]
\caption{Quantum Cognition Machine Learning intrinsic dimension estimator}\label{alg:main_algo}
\KwData{Data set $X \subseteq \mathbb{R}^D$ lying on a data manifold $M \subseteq \mathbb{R}^D$}
\KwResult{A list \texttt{dlist} of local intrinsic dimension estimates $d_x \approx \dim_x M$ }

Train a matrix configuration $A = \{A_1, \ldots , A_D\}$ on $X$ as in \eqref{eqn:matrix_config_optimization} or \eqref{eqn:matrix_config_optimization_with_variance}\;
\texttt{dlist} $\gets \emptyset$ \;
\For{$x\in X$}{
 calculate the ground state $\ket{\psi_0(x)}$ of the error Hamiltonian $H(x)$ \;
 calculate the position $y = A(\psi_0(x)) \in X_A$  \;
 calculate the spectrum $e_0 \leq  \ldots \leq e_{D-1}$ of the quantum metric $g(y)$ \;
 Calculate $\gamma = \mathrm{argmax}_{i=1, \ldots, D}\;  e_i/e_{i-1}$, the largest spectral gap \;
 Append $d = D - \gamma$ to  \texttt{dlist}
}

Return \texttt{dlist}
\end{algorithm}

 The Algorithm \ref{alg:main_algo} returns a list of intrinsic dimension estimates for every point $x \in X$. To extract a global estimate, a variety of techniques can be employed, such as taking the mode, median, or geometric mean to more refined $k$-nearest neighbor techniques. A global estimate can also be easily adapted to the case where multiple connected components of $M$ are detected, each with possibly different dimensions. Note that in steps 7-8 of Algorithm \ref{alg:main_algo}, we calculate the largest spectral gap by comparing successive ratios of eigenvalues. With this approach, the results $d=0,D$ cannot be detected. We are indeed assuming through the article that the data manifold does not have zero-dimensional/codimesion zero connected components.  
 
 It is possible to replace these crude spectral gap estimates with more advanced methods. For example, if $D$ is large and $d \ll D$, as is typical in real data sets, methods based on random matrix theory\cite{donoho1} are likely to give more robust estimates. Phase transitions in random matrix theory (RMT) refer to the abrupt changes in the behavior of eigenvalues of large random matrices as certain parameters are varied. These transitions are particularly interesting because they often separate different regimes of matrix behavior.  The eigenvalues of large random matrices follow well-defined distributions (like the Marchenko-Pastur distribution \cite{marchenko1967}) and as the matrix size grows, eigenvalue behavior exhibits certain regularities, with interesting gaps between signal and noise eigenvalues.  There is often a critical threshold phase transitions at which the behavior of the eigenvalues changes sharply. The presence of spectral gaps between eigenvalues can signal the existence of significant phase transition and in high-dimensional problems, RMT can predict the existence of these gaps. Furthermore, the eigenvectors associated with eigenvalues that exhibit an eigen-gap will be informative and uninformative when the eigen-gap vanishes \cite{nadakuditi2013most}.

One approach to recover the true signal matrix is to threshold the singular values of the quantum metric $g(y)$ and keep the singular values that are likely to correspond to the signal and discard those that are likely to be noise \cite{donoho1}. This leads to a singular value thresholding rule, where a threshold $\tau$ is applied to the singular values of the observable matrix, and only the singular values larger than $\tau$ are retained.  It was shown that in the asymptotic limit as $t, D \rightarrow \infty$ with $t/D\rightarrow \gamma$,  the optimal threshold is $\tau_{opt}=\frac{4}{\sqrt{3}} \cdot \sigma$ where $\sigma$ is the standard deviation of an underlying Gaussian noise matrix \cite{donoho1}.  The noise parameter $\sigma$ can be estimated by $\hat{\sigma}$ using the Marchenko-Pastur bulk singular values. This estimate can then be used to adaptively set the threshold for singular value thresholding. Specifically, the  rule  $\hat{\tau}_{opt}=\frac{4}{\sqrt{3}}\cdot \hat{\sigma}$ can be applied to the singular values of the of the quantum metric $g(y)$ for hard thresholding to find the spectral gap. In the following, we will refer to this thresholding method as the ``RMT-based" estimate.

The choice of dimension $N$ of the Hilbert space underlying the matrix configuration $A$ is a hyperparameter of the algorithm. As shown in \cite{Schneiderbauer_2016, oxidation}, we have the rank bound
\begin{equation}
\label{eqn:rank_bound}
\rk g \leq 2(N-1),
\end{equation}
so that $N$ should be chosen large enough to ensure $2(N-1) > d$. Since a priori we only know that $d<D$, a sensible choice would be to set $N \geq D/2 + 1$. However, for large datasets with $D \gg 0$, this choice might be impractical, since the number of parameters of a QCML estimator scales quadratically in $N$. Instead, a simple strategy for choosing $N$ that we employ in large real data sets is to first pick $N$ small and gradually increase it until a clear spectral gap emerges and is consistent across different choices of dimensions. In general, larger Hilbert space dimension $N$ will result in point clouds $X_A$ that are closer to the original data $X$ (low bias) but may also model noise artifacts (high variance). A smaller $N$ will result in approximations that may have higher loss/higher energy (high bias) but that may be more robust with respect to noise (low variance). 
 
\subsection*{Benchmarks}

\subsubsection*{The fuzzy sphere}
\begin{figure}[H]
\centering

\begin{subfigure}{0.45\textwidth}
    \centering
    \includegraphics[scale=0.2]{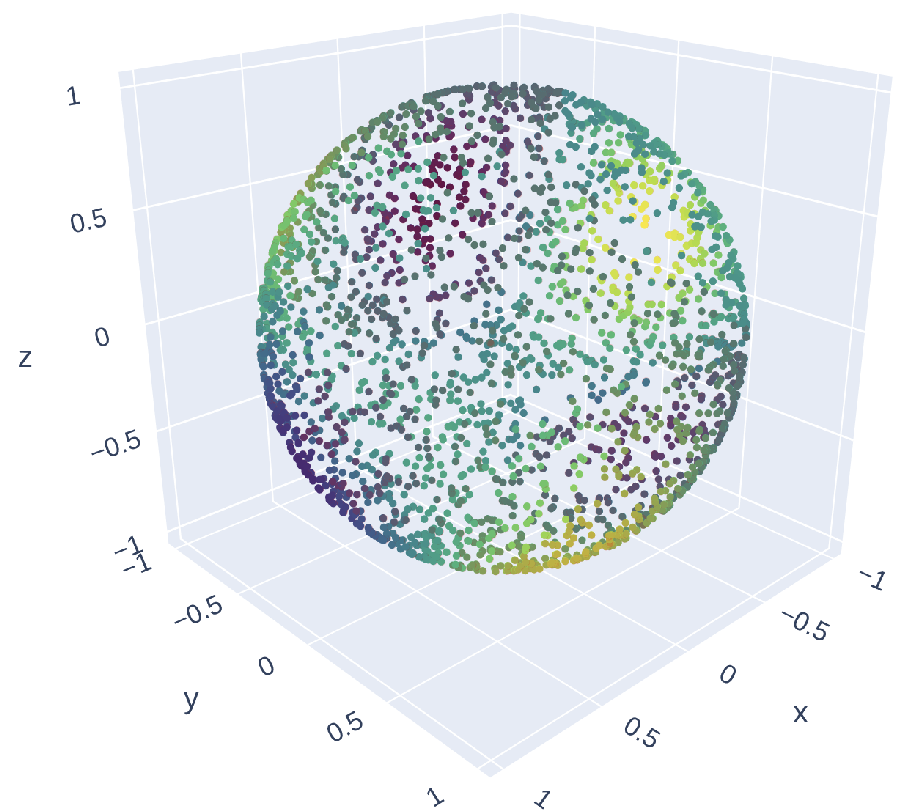}
    \caption{} 
\end{subfigure}
\begin{subfigure}{0.45\textwidth}
    \centering
    \includegraphics[scale=0.11]{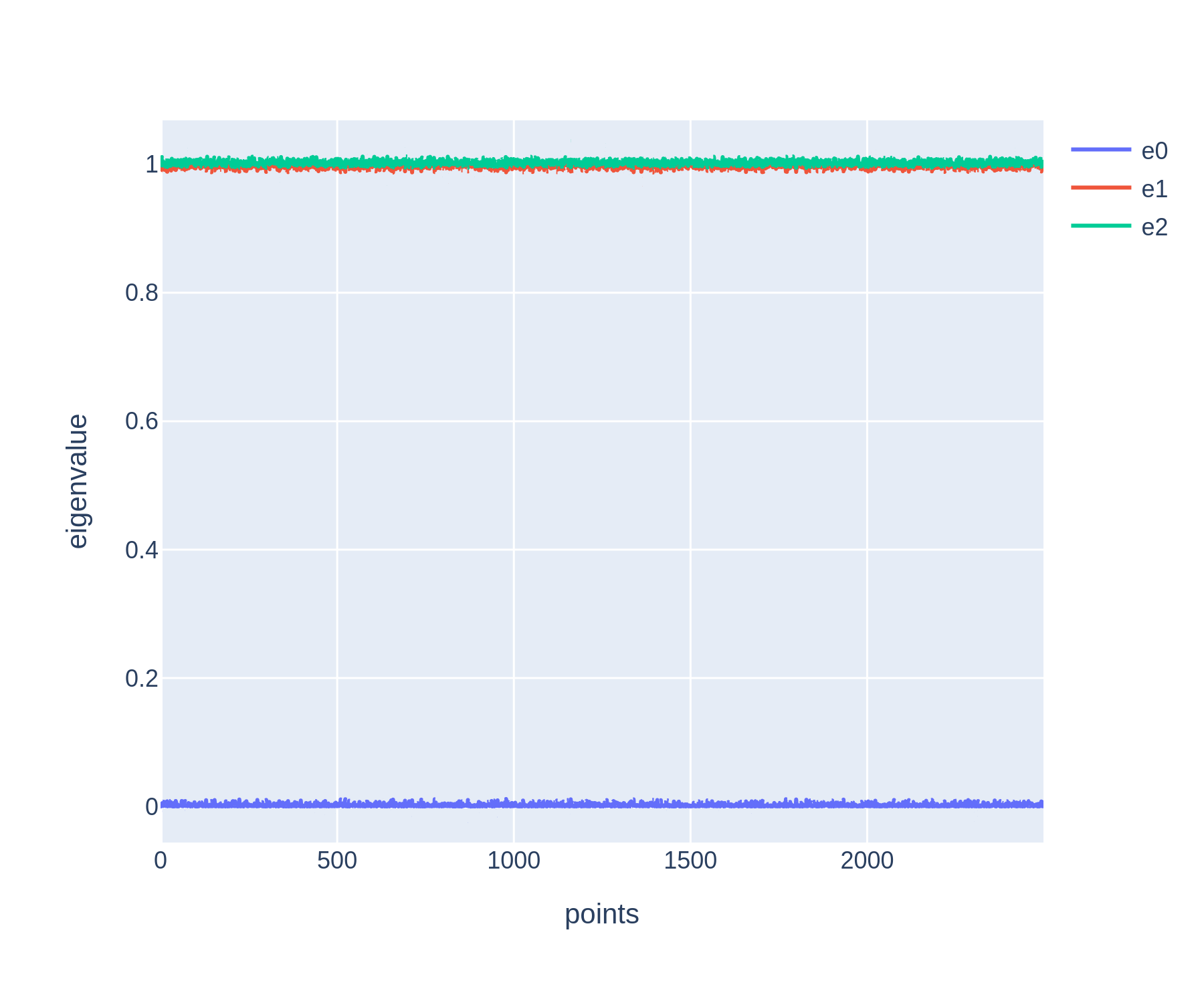}
    \caption{}
\end{subfigure}

\medskip

\begin{subfigure}{0.45\textwidth}
    \centering
    \includegraphics[scale=0.2]{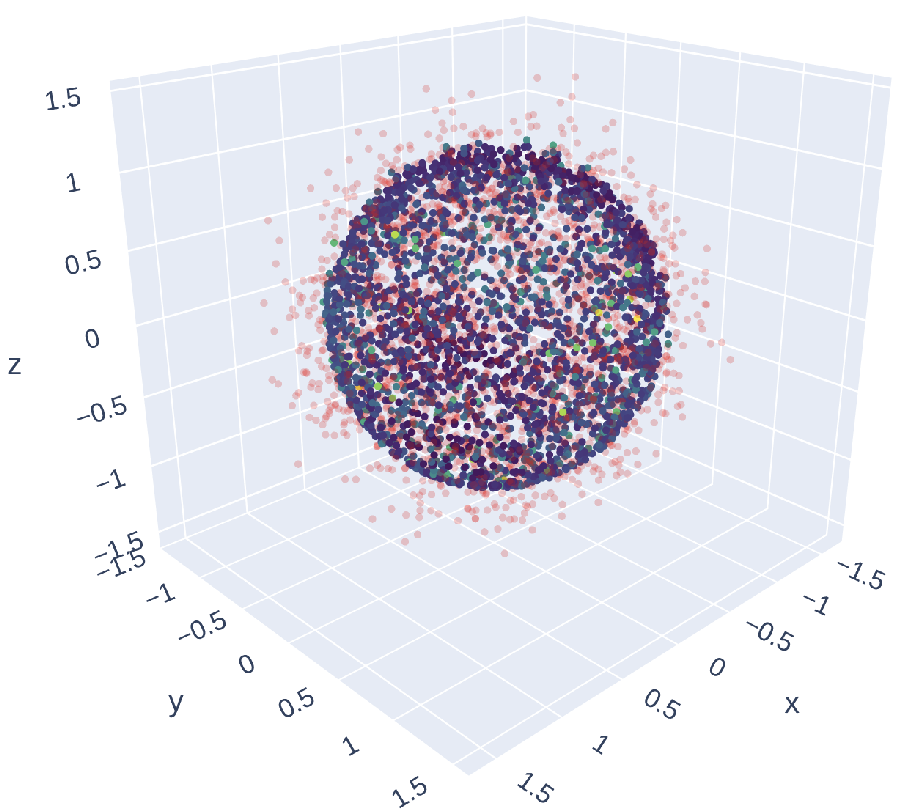}
    \caption{} 
\end{subfigure}
\begin{subfigure}{0.45\textwidth}
    \centering
    \includegraphics[scale=0.11]{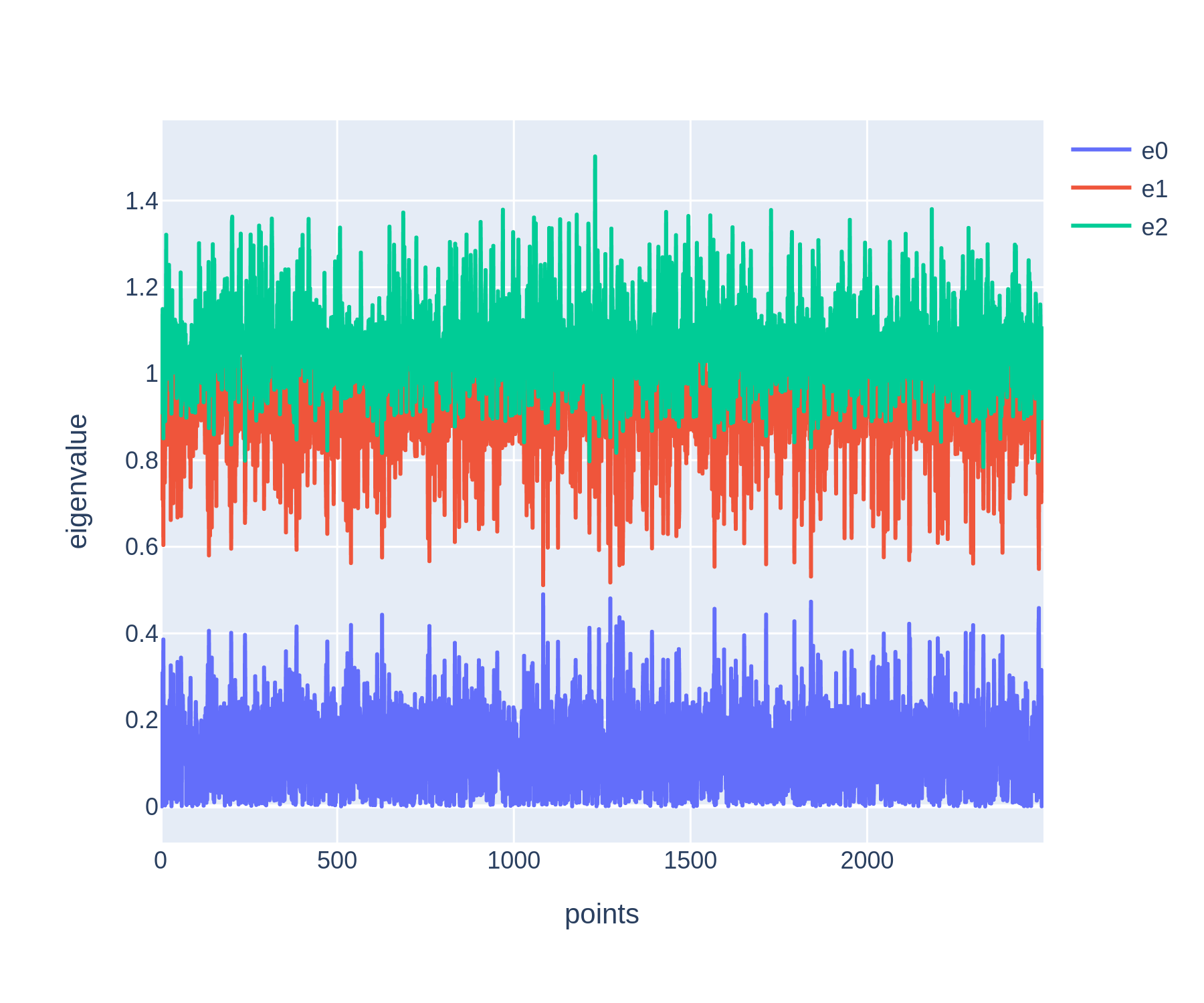}
    \caption{}
\end{subfigure}
\caption{Two configurations are shown for a data set $X$ consisting of $T=2500$ points uniformly distributed on the unit sphere with different levels of noise. (a,c) Scatter plot of the point cloud $X_A$ for (a) \texttt{noise = 0}, and (c) \texttt{noise = 0.2}, for two corresponding matrix configurations $A$ trained with Hilbert space dimension $N=3$. The original dataset is overlayed in red. Darker points correspond to lower relative error energy $E_0$. (b,d) Spectral gaps for (b) \texttt{noise = 0} and (d) \texttt{noise = 0.2}. The $x$-axis corresponds to points $x\in X_A$ and on the $y$-axis the eigenvalues of the quantum metric $g(x)$ are plotted. }
\label{fig:unit_sphere_results}
\end{figure}

We first evaluate Algorithm \ref{alg:main_algo} in the case when the data $X$ is a sample of $T=2500$ uniformly distributed points on the unit sphere $M = S^2$, embedded in $D=3$ dimensions. We allow the data to be ``noisy'', that is, $x \in X$ might not necessarily lie on $M$ but it could be drawn from a Gaussian distribution whose mean is on $M$ and whose standard deviation is a \texttt{noise} parameter. By the rank bound in \eqref{eqn:rank_bound} on the quantum metric, the minimum possible choice of Hilbert space dimension is $N=3$. Plots of the point cloud $X_A$ and the spectra of the quantum metric $g(x)$ at different points  $x\in X_A$ are shown in Figure \ref{fig:unit_sphere_results}. With zero noise (Figure \ref{fig:unit_sphere_results} a-b) the point cloud approximation $X_A$ is very close to the original unit sphere and a clear spectral gap emerges at every point between the top 2 eigenvalues of the quantum metric and the lowest eigenvalue. The intrinsic dimension estimate is thus $d=2$ at all points. As the noise level increases, up to $\texttt{noise}=0.2$ (Figure \ref{fig:unit_sphere_results} c-d) the point cloud starts picking up some noise artifacts and the variance of the metric spectrum increases. However, even for $\texttt{noise}=0.2$, the intrinsic dimension estimate is $d=2$ at 2471 points out of 2500, giving an average estimate of $d\approx 1.9884$. 

\newpage

\begin{figure}[H]
\centering
\begin{subfigure}{0.32\textwidth}
    \centering
    \includegraphics[scale=0.1]{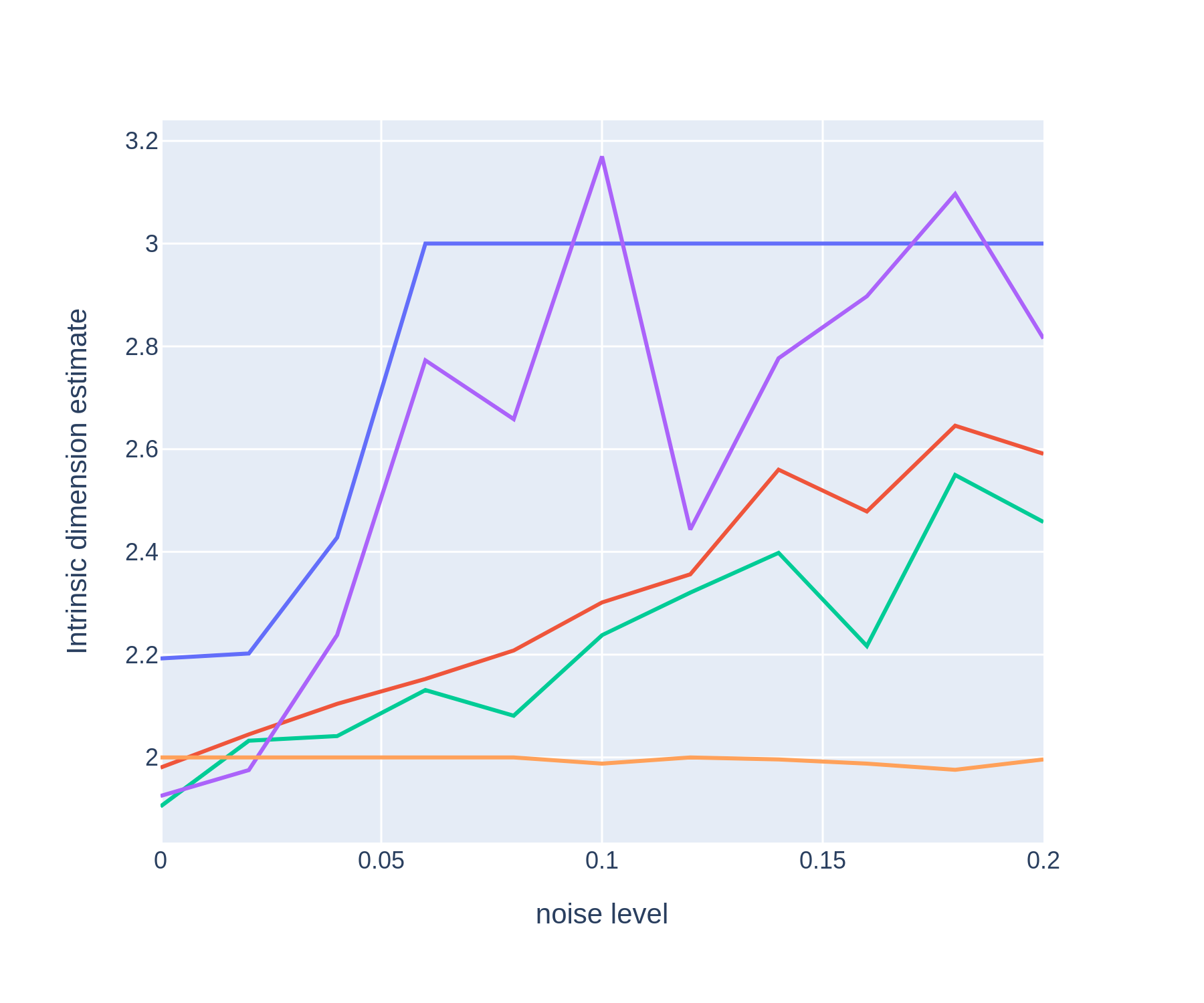}
    \caption{}
\end{subfigure}
\begin{subfigure}{0.32\textwidth}
    \centering
    \includegraphics[scale=0.1]{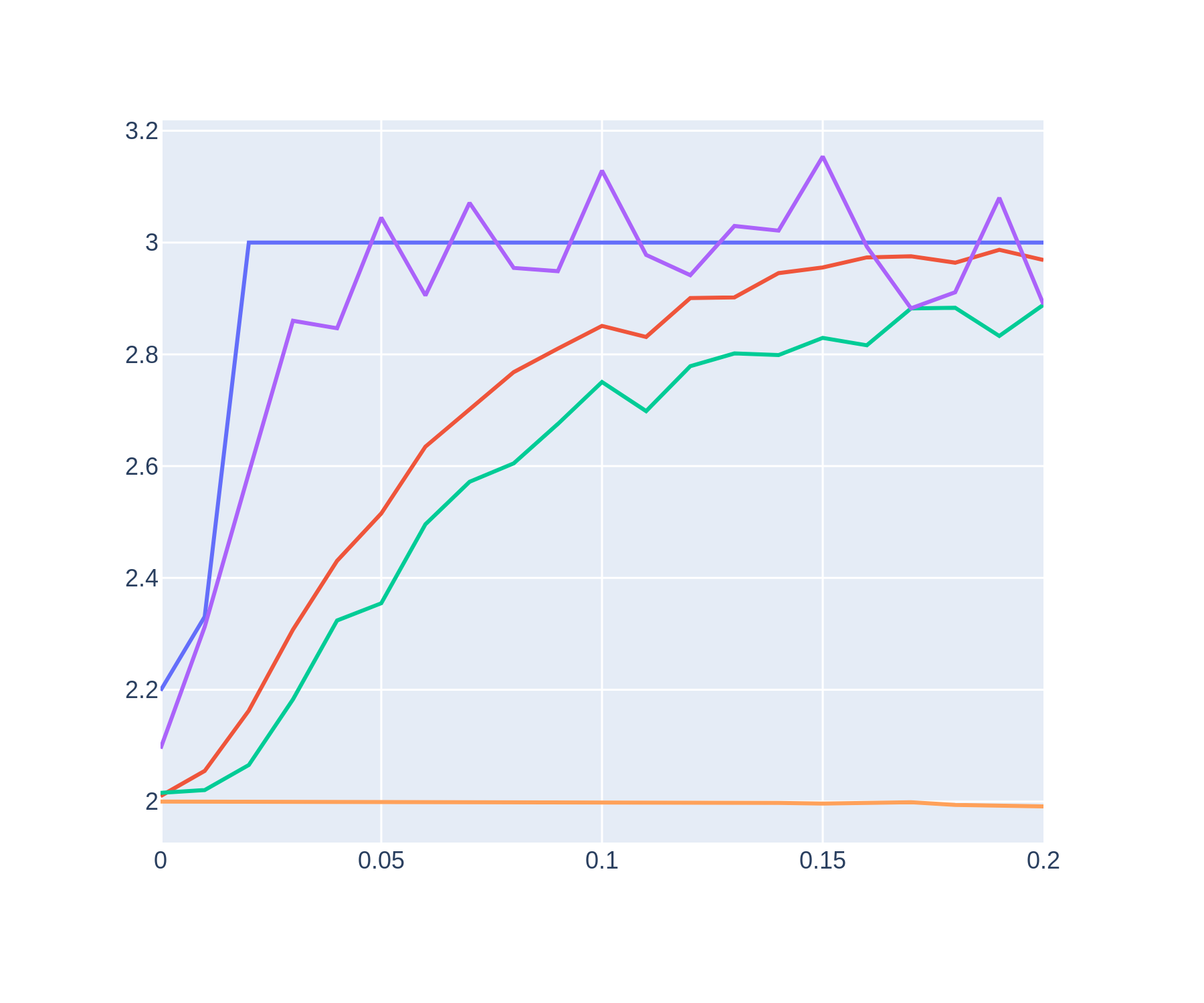}
    \caption{}
\end{subfigure}
\begin{subfigure}{0.32\textwidth}
    \centering
    \includegraphics[scale=0.1]{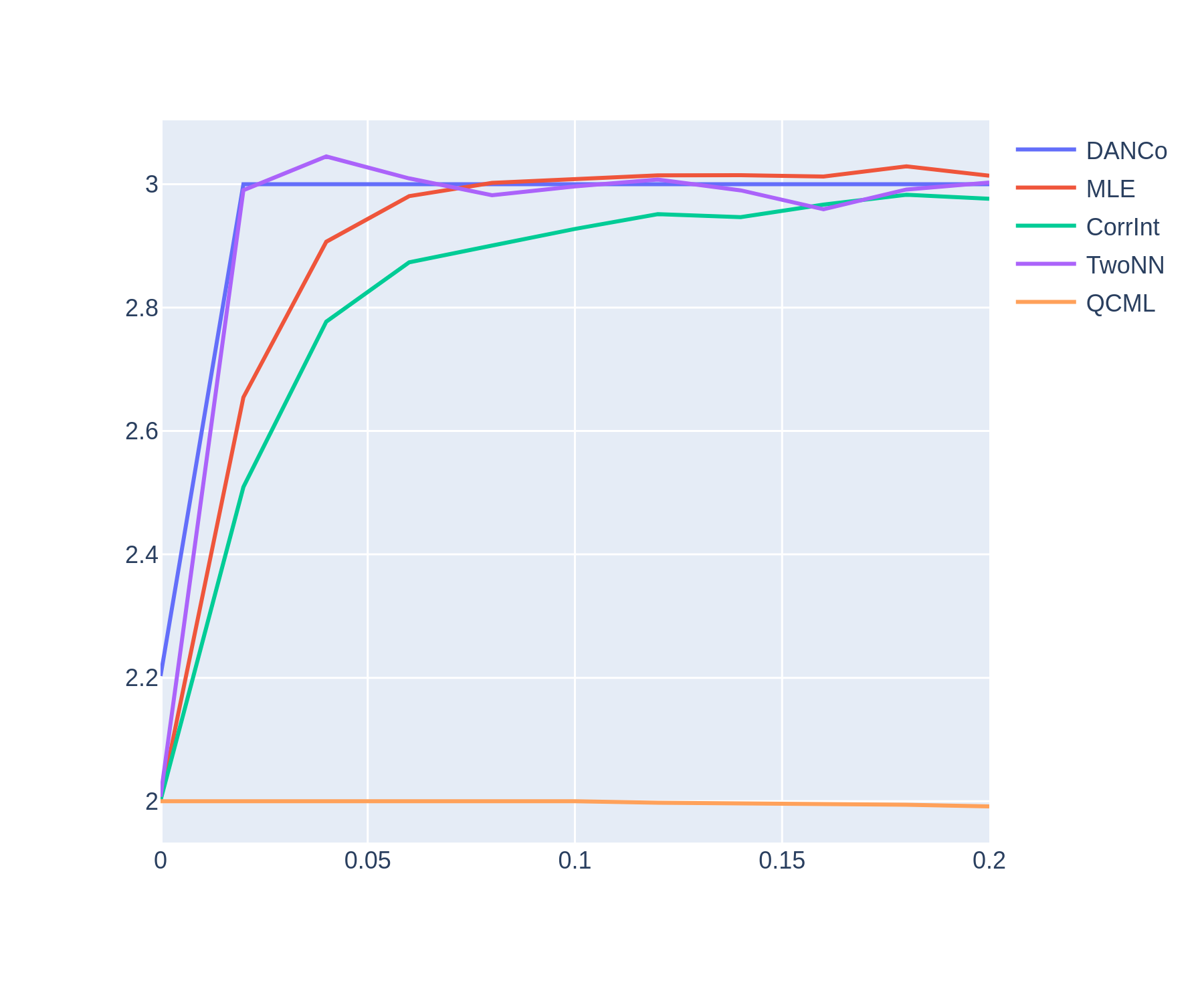}
    \caption{}
\end{subfigure}
\caption{Intrinsic dimension estimates for the unit sphere $S^2$ as a function of \texttt{noise} level. Varying data set sizes of (a) $T=250$, (b) $T=2500$, (c) $T=25000$ points are tested. For the QCML estimator, the average estimate across all $T$ points is plotted. A slight degradation in the estimate for the QCML estimator is noticeable for \texttt{noise} > 0.15, especially in the case of $T=250$, but it is otherwise robust when compared to other methods.}
\label{fig:fuzzy_sphere_benchmarks}
\end{figure}

For comparison, we selected some of the best-performing state-of-the-art algorithms for intrinsic dimension estimation (DANCo,  MLE, CorrInt, MiND ML, TwoNN, as implemented in ref. \cite{scikit-dimension} ) and tested them at different levels of noise, and for three different data set sizes $T=250, 2500, 25000$ (Figure \ref{fig:fuzzy_sphere_benchmarks}). In Figure \ref{fig:fuzzy_sphere_benchmarks}, the slope of the intrinsic dimension estimate for the QCML model is nearly zero, so that the estimate is essentially unaffected by noise in the range 0-0.2. The dimension estimate is also consistent across different number of samples $T=250, 2500, 25000$, indicating additional robustness with respect to data distribution and density. By comparison, the estimates of all other baseline algorithms quickly converge to $d=3$, creating a ``shadow'' dimension out of the noise. Increasing the size of the sample does not seem to aid the state-of-the-art algorithms in detecting the correct intrinsic dimension. In fact, the slopes of the ``shadow dimension'' graphs in Figure \ref{fig:fuzzy_sphere_benchmarks} get noticeably steeper for $T=25000$ samples, indicating an even faster degradation of the intrinsic dimension estimate as the data density increases. 

It is perhaps worth noting that the optimal matrix configuration $A = \{A_1, A_2, A_3\}$ obtained by the QCML estimator in this case are well-known to physicists as ``fuzzy spheres'' \cite{Ishiki, Schneiderbauer_2016, Steinacker_2021}. Up to a change of basis and a re-scaling factor, the elements of $A$ are given by the angular momentum operators in quantum mechanics.  

\subsubsection*{Higher-dimensional synthetic manifolds}

\begin{figure}
\centering

\begin{subfigure}{0.32\textwidth}
    \centering
    \includegraphics[scale=0.1]{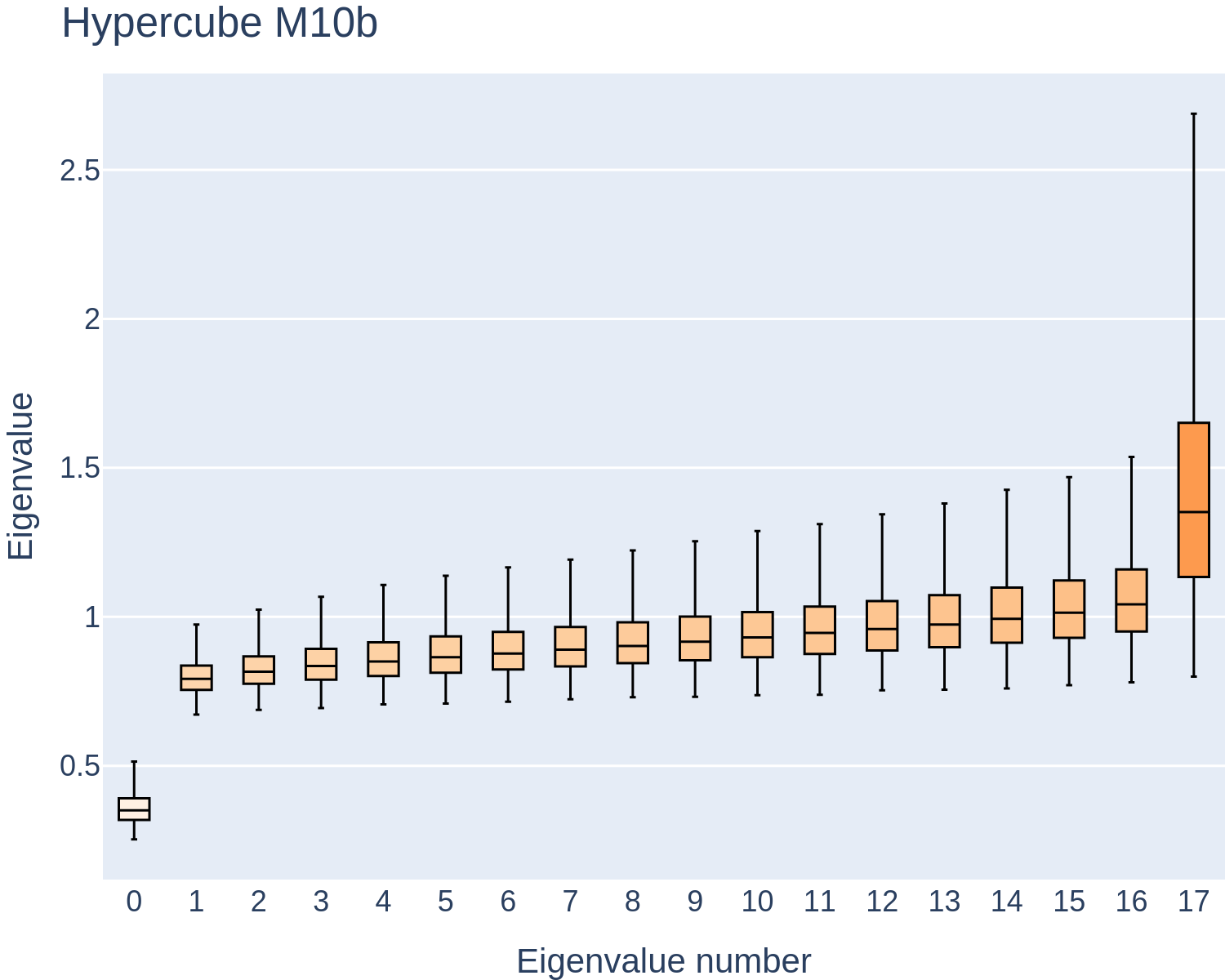}
    \caption{} 
\end{subfigure}
\begin{subfigure}{0.32\textwidth}
    \centering
    \includegraphics[scale=0.1]{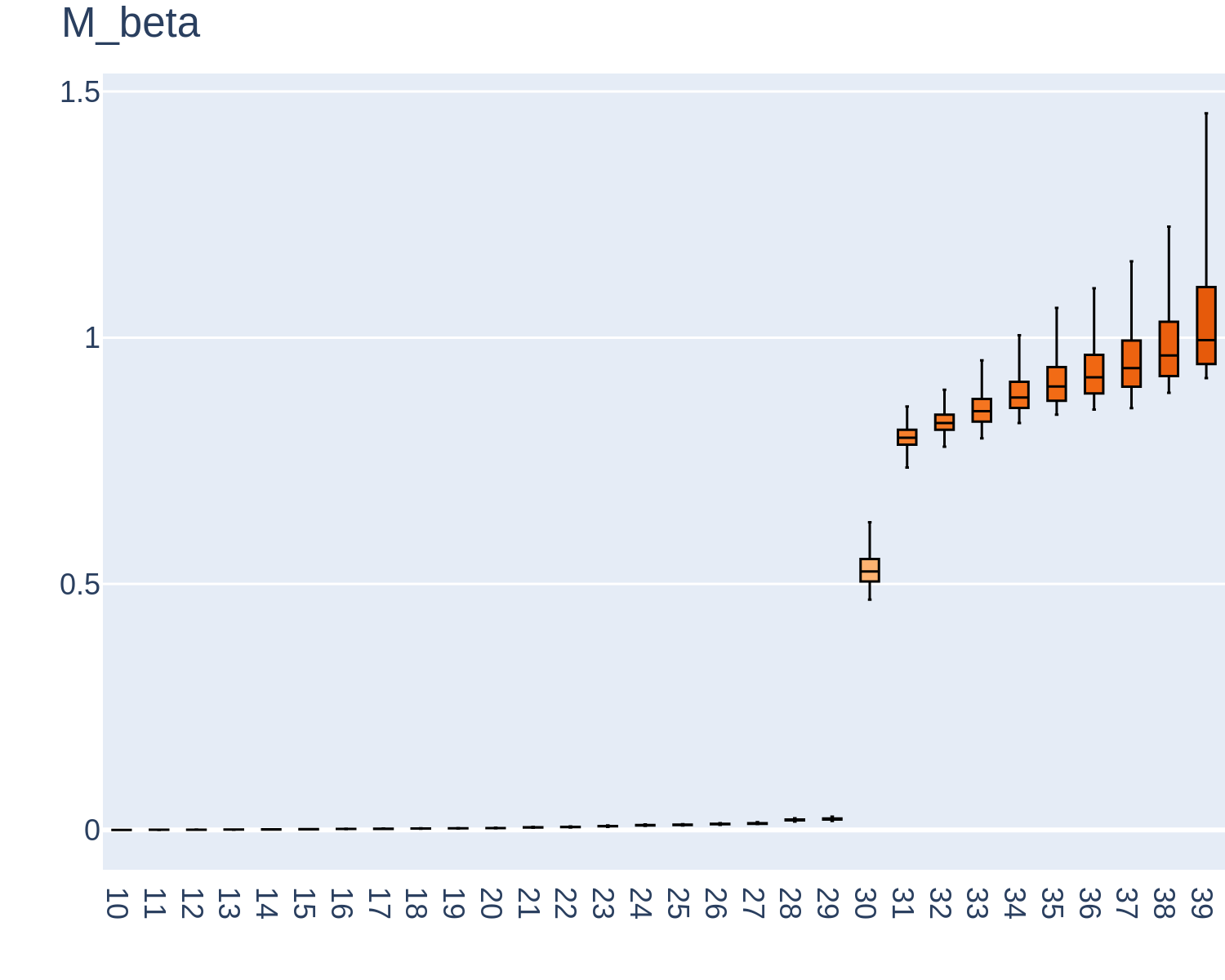}
    \caption{}
\end{subfigure}
\begin{subfigure}{0.32\textwidth}
    \centering
    \includegraphics[scale=0.1]{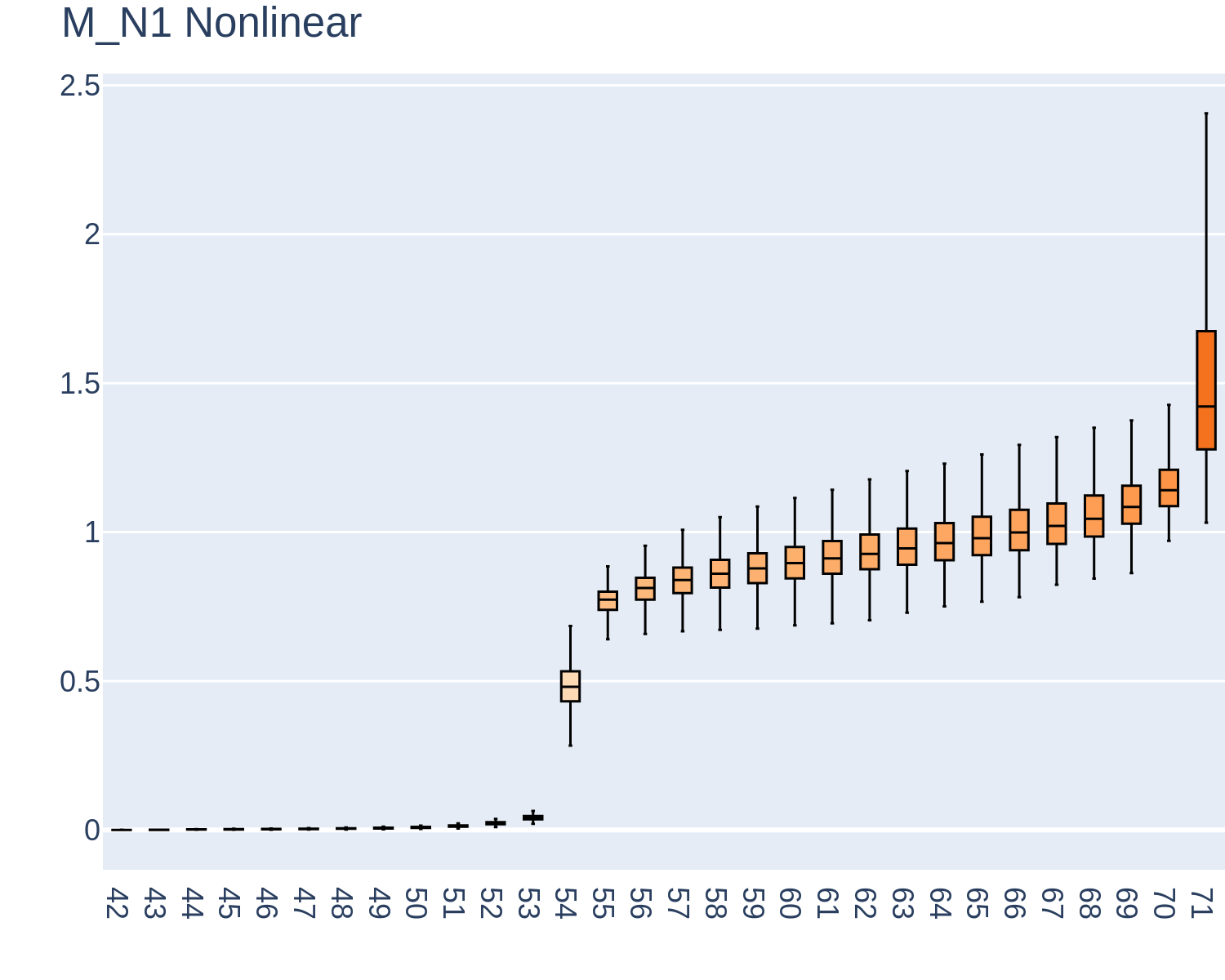}
    \caption{}
\end{subfigure}

\medskip

\begin{subfigure}{0.32\textwidth}
    \centering
    \includegraphics[scale=0.1]{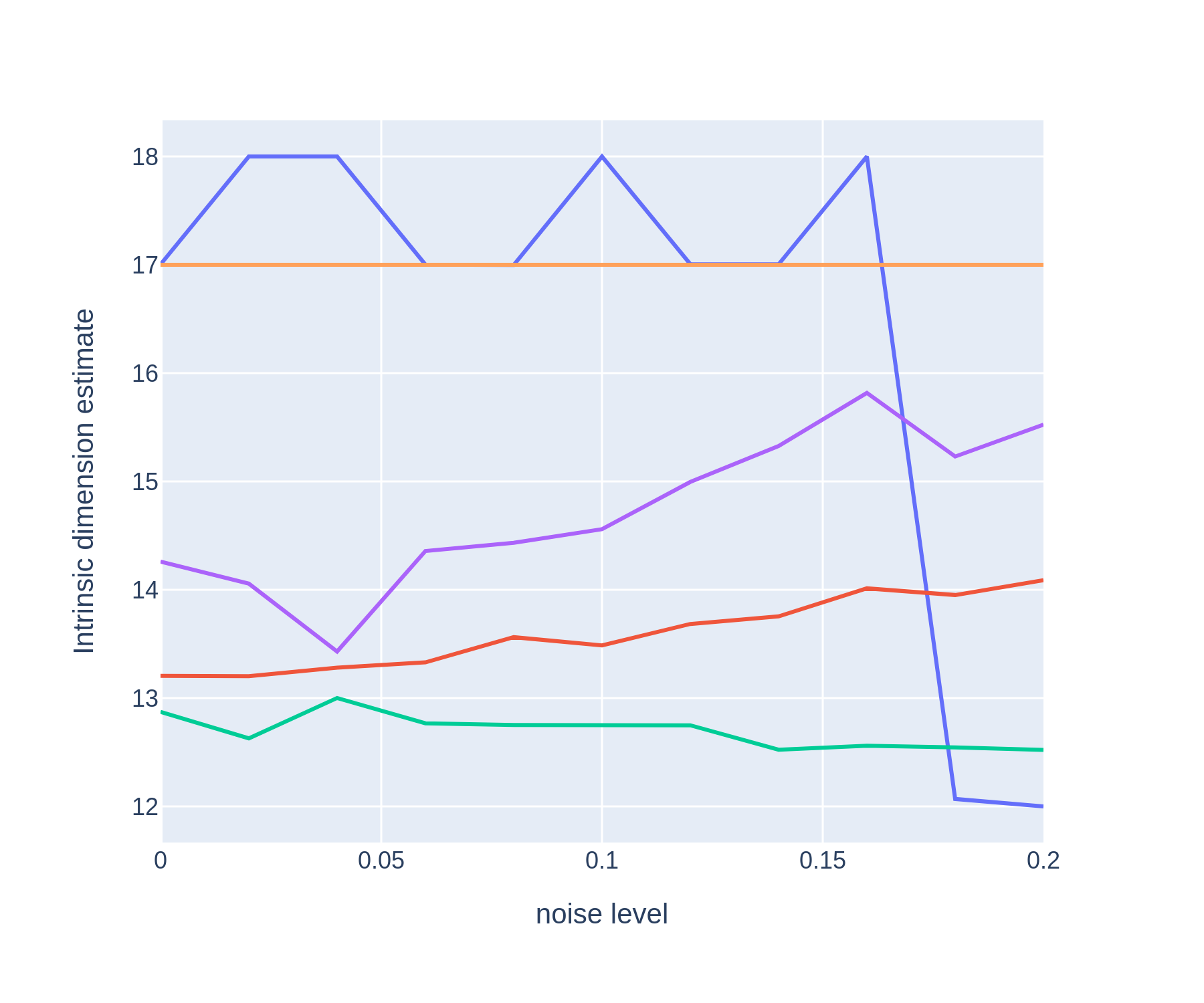}
    \caption{} 
\end{subfigure}
\begin{subfigure}{0.32\textwidth}
    \centering
    \includegraphics[scale=0.1]{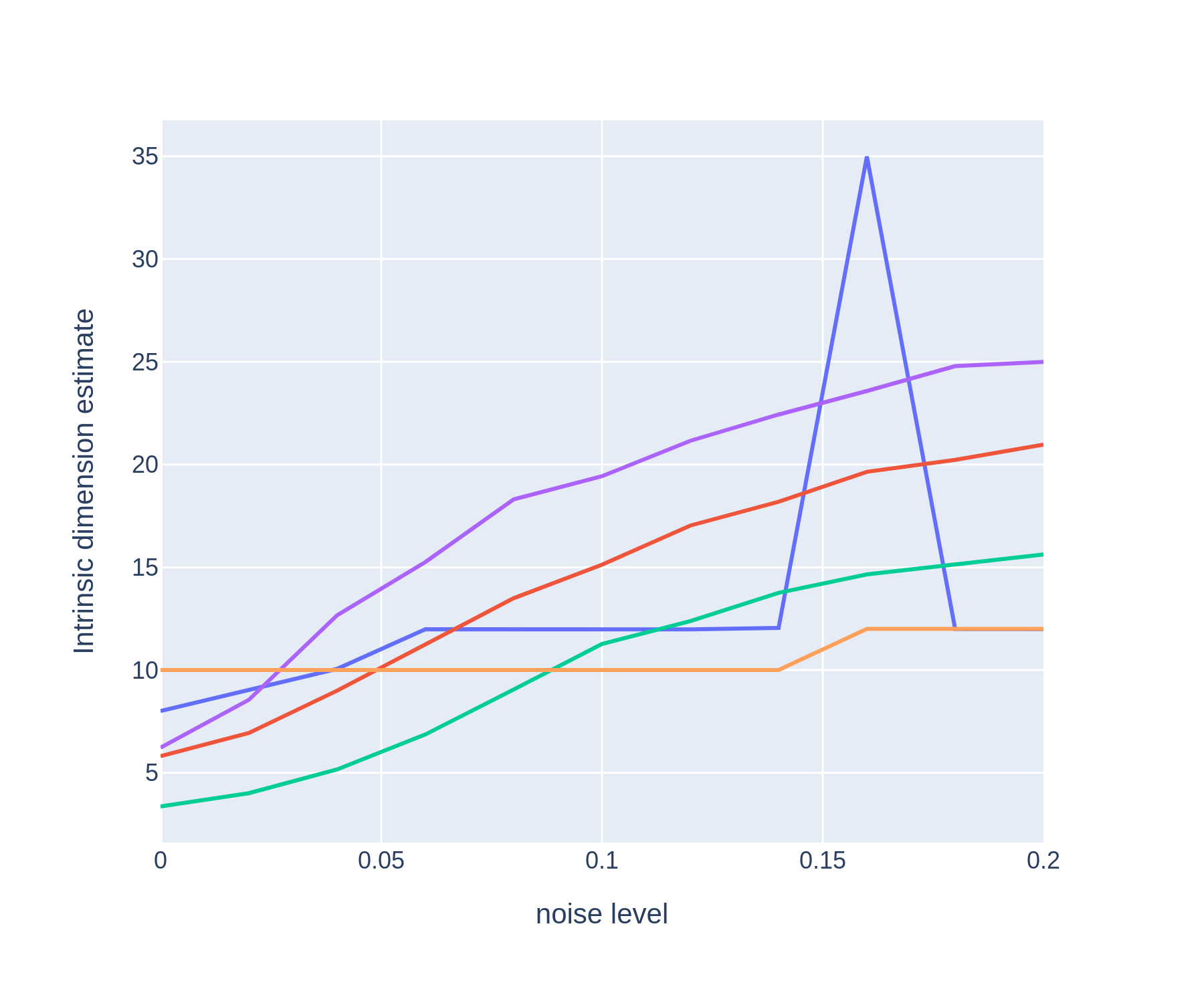}
    \caption{}
\end{subfigure}
\begin{subfigure}{0.32\textwidth}
    \centering
    \includegraphics[scale=0.1]{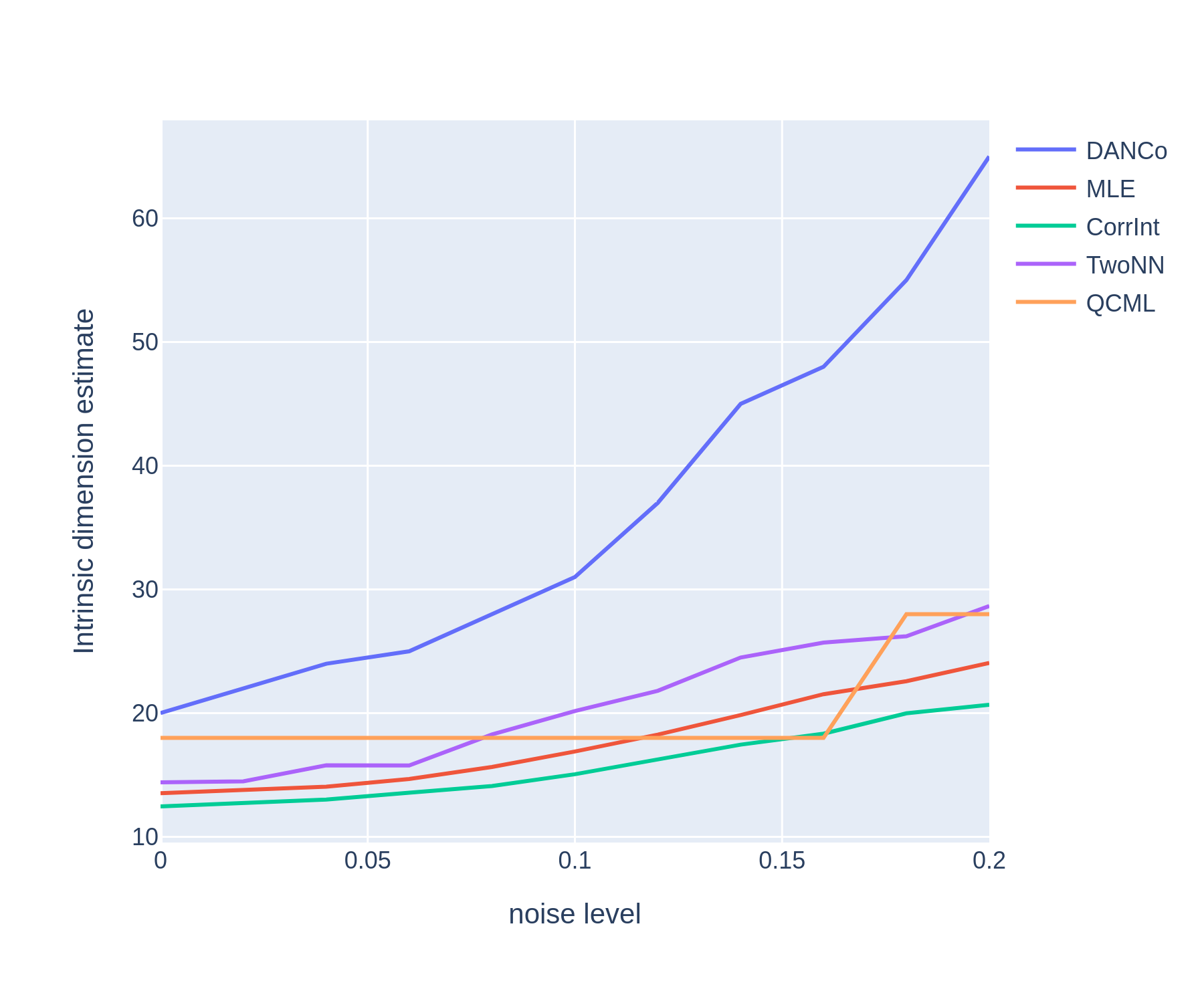}
    \caption{}
\end{subfigure}

\caption{Intrinsic dimension estimates for $T=2500$ points on three higher-dimensional benchmark manifolds\cite{Campadelli}: the 17-dimensional hypercube $M10b$, the 10-dimensional $M_{\beta}$ manifold embedded into $D=40$ dimensions, and the 18-dimensional manifold $MN_1$ embedded non-linearly into $D=72$ dimensions. In the boxplots (a-c) the $i$-th box represents the distribution of the eigenvalue $e_i$ across all $T=2500$ points. The outliers have been omitted from the plot for clarity. The plots (d-f) show the intrinsic dimension estimates for each manifold as functions of the \texttt{noise} parameter. In these examples a global estimate of dimension for the QCML estimator was obtained by taking the median of the local dimension estimates.}
\label{fig:higher_dim_benchmarks}
\end{figure}

Next, we test the QCML estimator \ref{alg:main_algo} on three higher-dimensional manifolds included in the benchmarking framework of ref. \cite{Campadelli}. The first is the 17-dimensional standard hypercube embedded into $D=18$ dimensions (Figure \ref{fig:higher_dim_benchmarks} (a), (d) ), and labeled $M10b$ in the $\texttt{scikit-dimension}$ Python package. The second is the 10-dimensional manifold $M_{\beta}$ (Figure \ref{fig:higher_dim_benchmarks} (b), (e) ), embedded in $D=40$ dimensions, and the third is the 18-dimensional manifold $MN_1$ (Figure \ref{fig:higher_dim_benchmarks} (c), (f) ) embedded into $D=72$ dimensions. These benchmarks are considered among the most difficult for intrinsic dimension estimation, due to both the non-uniform density of the data (for $M10b$ and $M_{\beta}$) and to the non-linearity of the embedding (for $M_{\beta}$ and $MN_1$). In our testing, we trained the QCML estimator with Hilbert space dimension $N=16$ on each of these manifolds, and plotted the distribution of the eigenvalues of the quantum metric $g(x)$ across all data points $x \in X$ (Figure \ref{fig:higher_dim_benchmarks} (a-c) ). In all cases, a clear spectral gap emerges between the top $d$ eigenvalues that are near 1, and the remaining $D-d$ bottom eigenvalues that are near 0, where $d$ is the correct intrinsic dimension.

These higher-dimensional manifolds can also be used as a testing ground for the random matrix theory (RMT) estimate of the spectral gap. Recall that this technique can be applied whenever the quantum metric is of low rank and of high dimension, and is therefore not suitable for the $S^2$ or $M10b$ examples. For $M_{\beta}$, the RMT estimate returns the correct dimension $d=10$. For $MN_1$, the artificial rank bound of 30 imposed by our choice of $N=16$ implies that the metric is not actually of low rank, and therefore the RMT estimate cannot be applied with this choice of $N$. We re-tested this example with a higher value $N=37$, the smallest dimension for which the rank bound is equal to the embedding dimension $D=72$, and obtained an estimate of $d=15$.

Next, we plotted the intrinsic dimension estimate returned by Algorithm \ref{alg:main_algo} as a function of Gaussian noise (Figure \ref{fig:higher_dim_benchmarks} (d-f) ) and compared the estimate to other standard intrinsic dimension estimators. A common theme among the standard estimators is to first under-estimate the intrinsic dimension, in the presence of zero or low noise. As explained in the introduction, this is a well-known effect due to the ``curse of dimensionality'', whereby neighboring points in high dimension tend to be very far apart. As the noise is increased, however, the ``shadow dimension'' effect overcomes the underestimating effects due to  sparsity and the standard algorithms begin to overestimate intrinsic dimension. This is particularly evident in the plots for $M_{\beta}$ and $MN_1$. By contrast, the spectral gap estimate of the QCML estimator is robust with respect to both these effects.

\subsubsection*{Image recognition data sets}

\begin{figure}[H]
\centering

\begin{subfigure}{0.49\textwidth}
    \centering
    \includegraphics[scale=0.1]{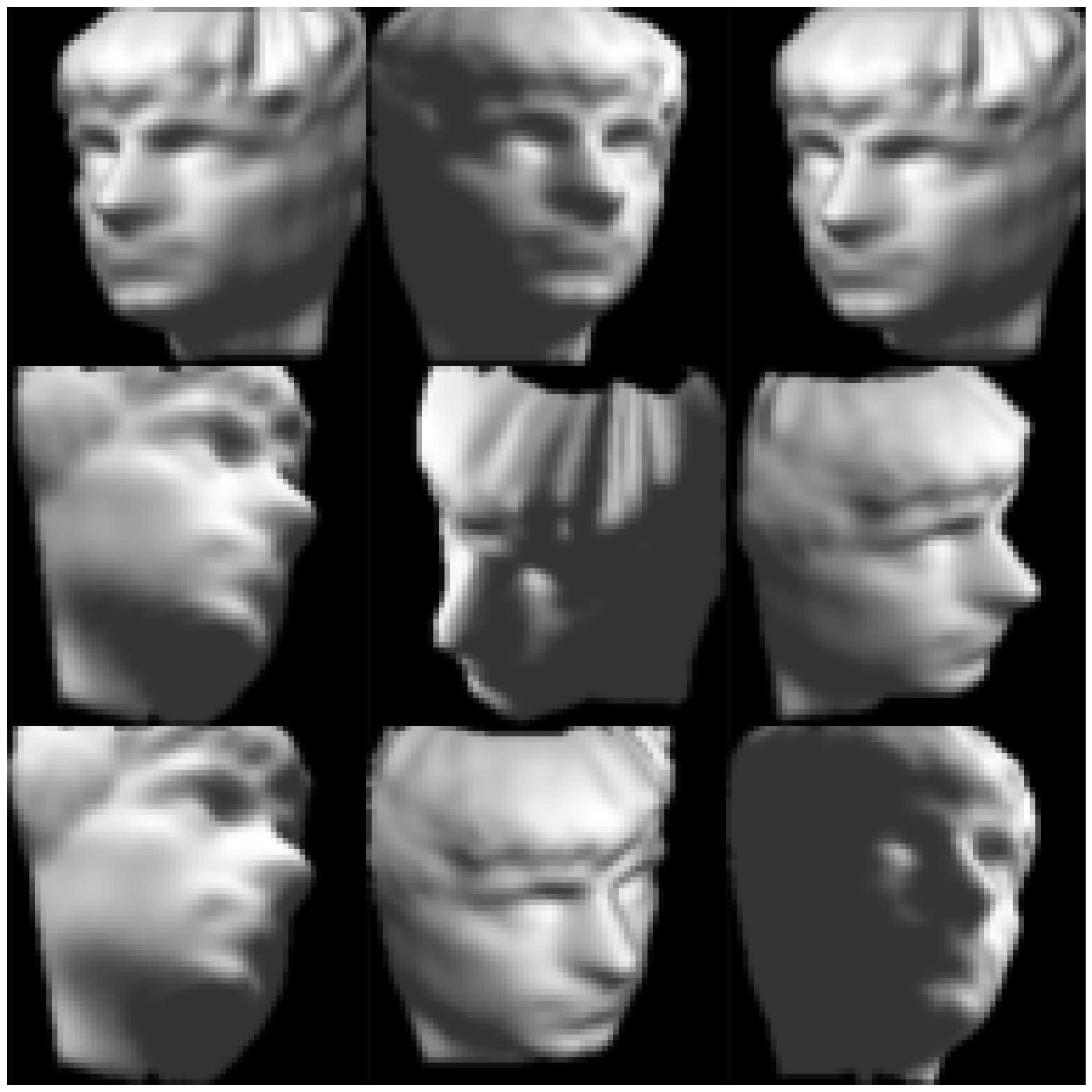}
    \caption{} 
\end{subfigure}
\begin{subfigure}{0.49\textwidth}
    \centering
    \includegraphics[scale=0.1]{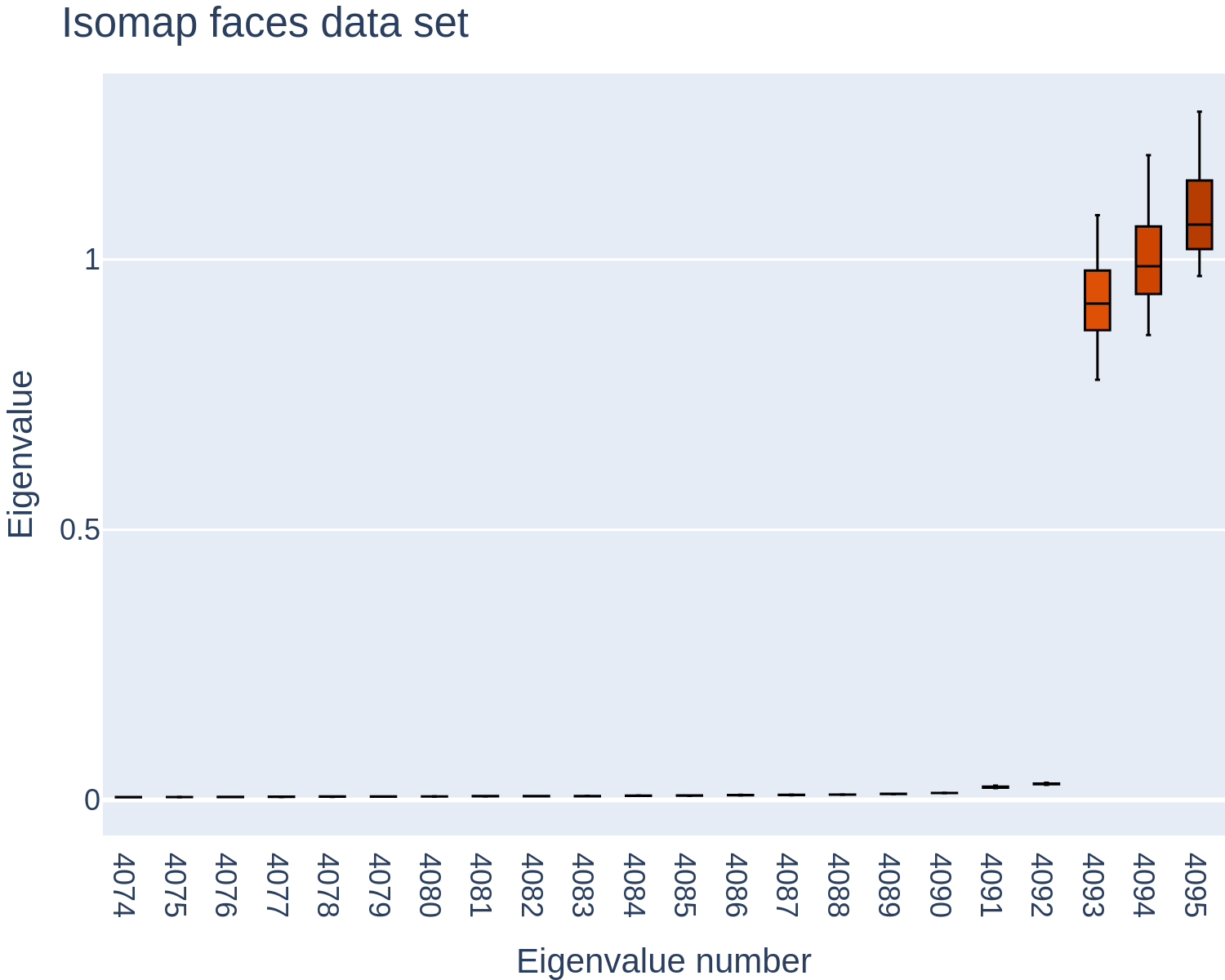}
    \caption{}
\end{subfigure}

\medskip

\begin{subfigure}{0.3\textwidth}
    \centering
    \includegraphics[scale=0.1]{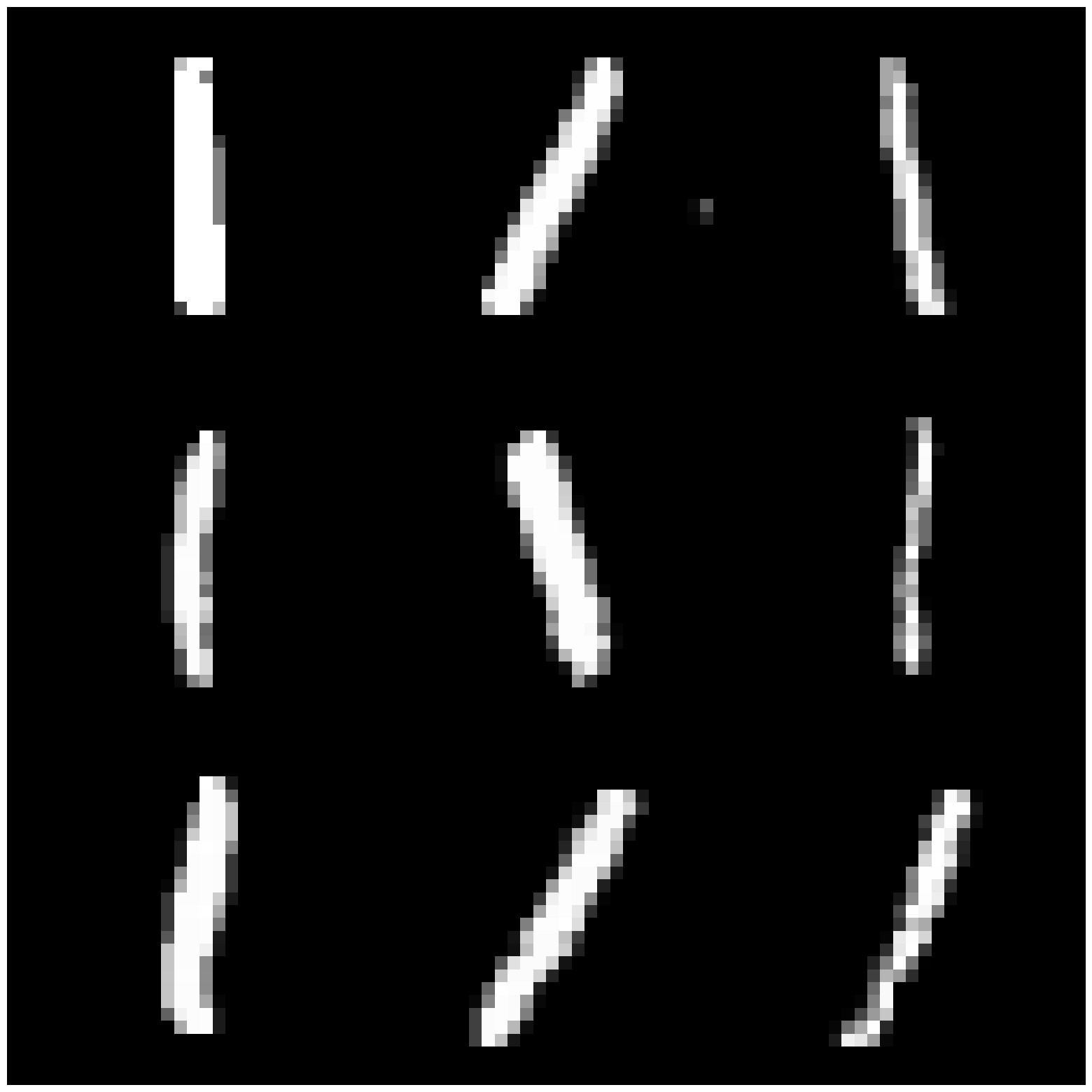}
    \caption{} 
\end{subfigure}
\begin{subfigure}{0.3\textwidth}
    \centering
    \includegraphics[scale=0.1]{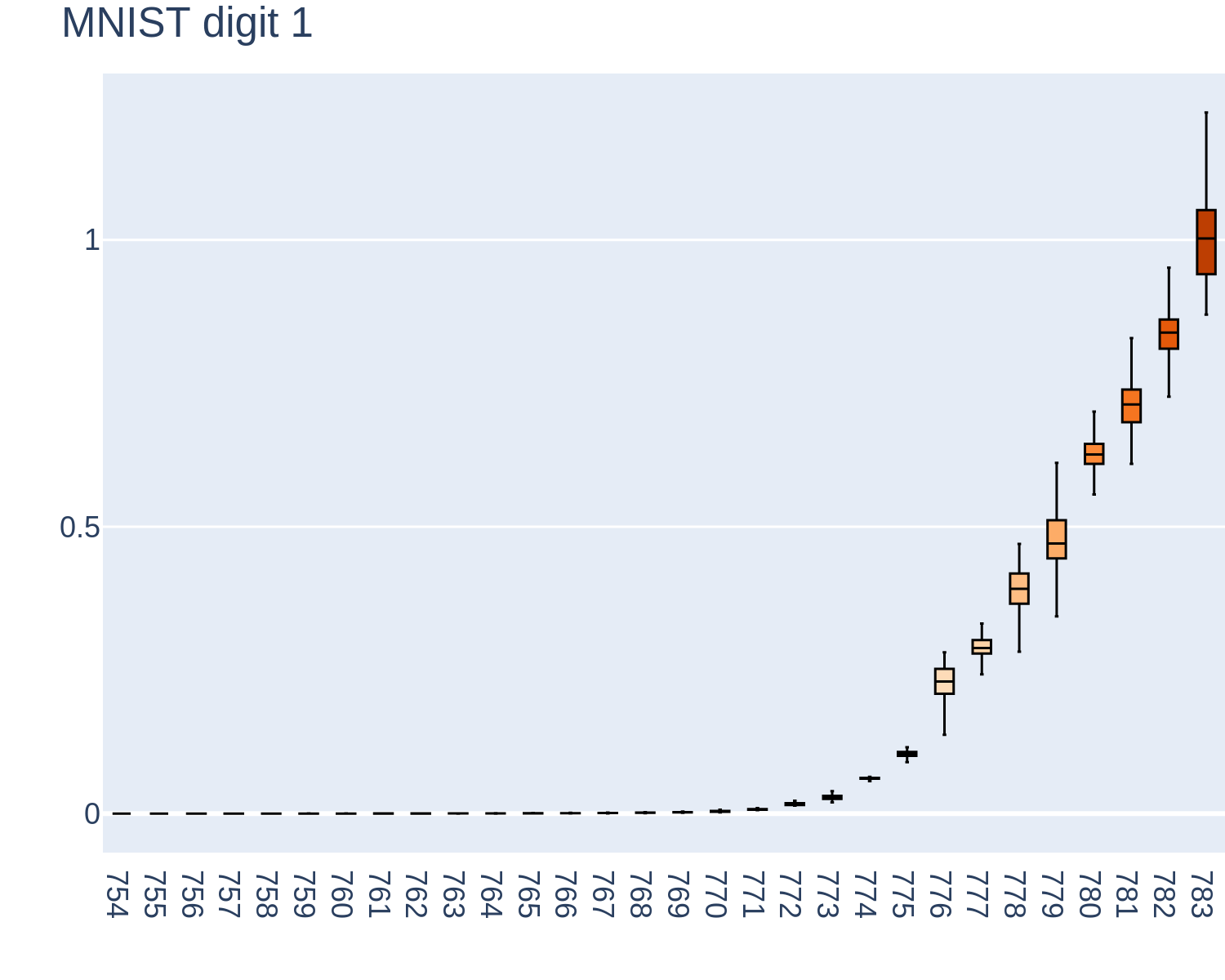}
    \caption{}
\end{subfigure}
\begin{subfigure}{0.3\textwidth}
    \centering
    \includegraphics[scale=0.08]{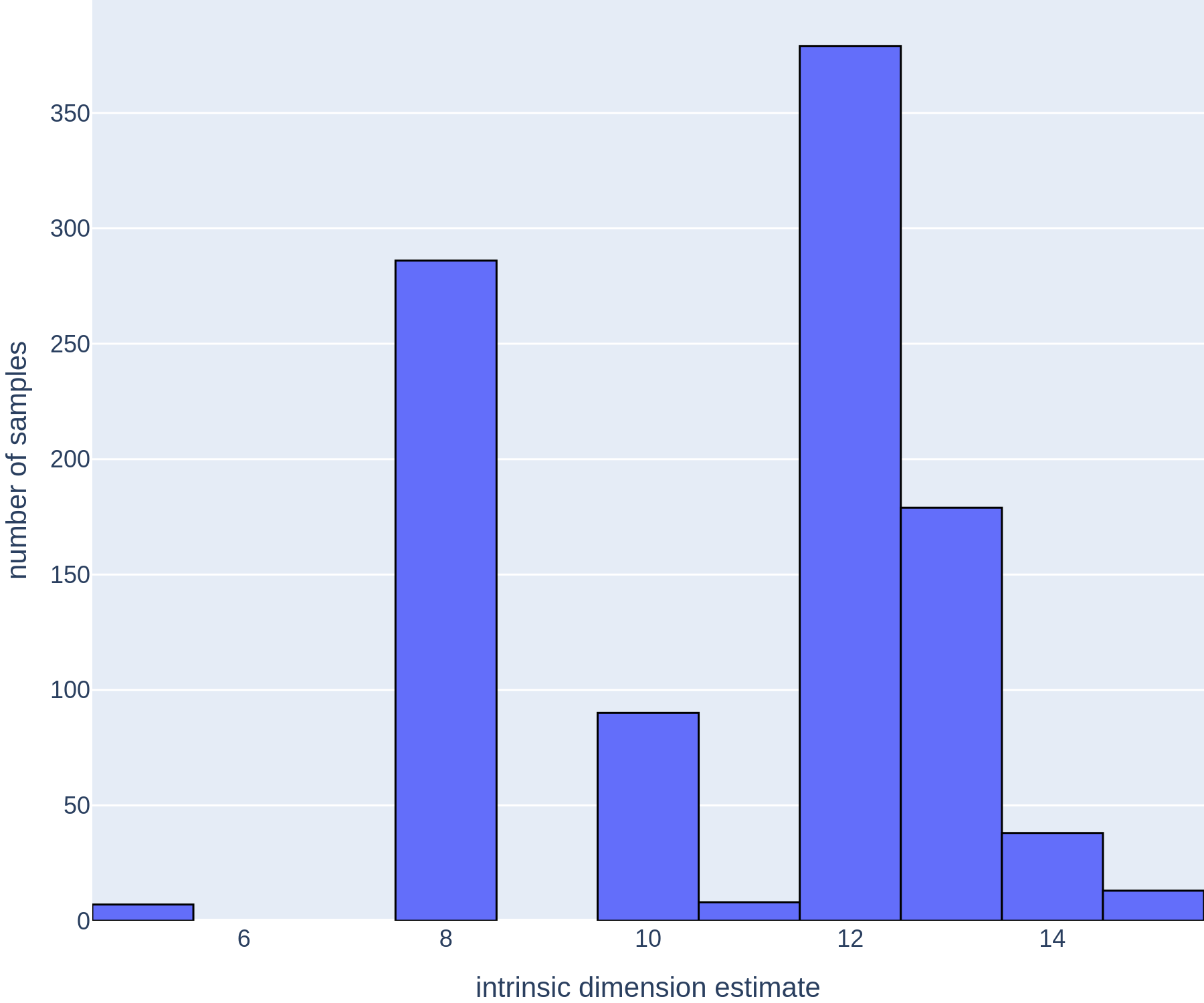}
    \caption{} 
\end{subfigure}

\caption{(a) Examples of images from the ISOMAP face database, (b) Spectral gap for ISOMAP, (c) Examples of digit ``1'' in the MNIST, (d) Spectral gap for MNIST digit ``1'', (e) Distribution of dimension estimates for MNIST digit ``1''. For ISOMAP, the distribution of dimension estimates is concentrated in dimension $d=3$, so the histogram is not shown.}
\label{fig:real_data}
\end{figure}

We next test the QCML estimator on two of the real data sets suggested in the benchmarking framework of ref. \cite{Campadelli}, the ISOMAP face database and MNIST. The ISOMAP face database consists of 698 grayscale images of size $64\times 64$ representing the face of a sculpture (Figure \ref{fig:real_data} (a)). Each image is represented as a vector in $D = 64^2 = 4096$ dimensions and it  corresponds to a different rotation with respect to two axes and a different lighting direction, so that the intrinsic dimension of the data manifold in this case is expected to be $d=3$. In Figure \ref{fig:real_data} (b) a well-defined spectral gap indeed emerges between the top 3 eigenvalues of the quantum metric and the remaining 4093. This result was obtained by training with Hilbert space dimension $N=32$. The value of $N=32$ was chosen after experimenting with different Hilbert space dimensions until a clear spectral gap emerged. The RMT-based intrinsic dimension estimate for ISOMAP faces is $d=3$.

The MNIST database consists of 70000 pictures of handwritten digits, each stored as a $28\times 28$ grayscale picture. The overall intrinsic dimension of this dataset is unknown, but it is expected that each digit has its own intrinsic dimension. For example, in ref. \cite{Hein_MNIST} estimates for the dimension of each digit are in the range $d = 8-14$. For our testing, we selected 1000 samples of the digit ``1'' (Figure \ref{fig:real_data} (c) ) and trained with Hilbert space dimension $N=16$. The results suggest that even within each digit there is variation of intrinsic dimension (Figure \ref{fig:real_data} (d-e)), perhaps reflecting different styles of handwriting. Indeed, in our testing we found a range of dimensions $d = 5-15$, with the majority of estimates being $d=8$ (286 samples) and $d=12$ (379 samples). This result highlights a further advantage of our method, which produces intrinsic dimension estimates for each data point independently, without requiring a sampling of its neighbors. The RMT-based intrinsic dimension estimate for MNIST digit ``1'' is $d=12$.

\subsubsection*{Wisconsin Breast Cancer data set }
\label{sec:WBC}

We also test our intrinsic dimension estimator on the Diagnostic Wisconsin Breast Cancer Database \cite{misc_breast_cancer_wisconsin_(diagnostic)_17}. This database consists of 569 data points representing images of fine needle aspirates (FNA) of a breast tumor. For each image, 30 features are extracted that describe characteristics of the cell nuclei present in the image. Therefore in this case we are sampling $T=569$ points from a manifold sitting inside $D=30$ dimensional Euclidean space. The dataset also contains labels indicating whether a certain tumor is benign or malignant, but for our intrinsic dimension testing the labels are discarded. 

\begin{figure}[H]
\centering

\begin{subfigure}{0.49\textwidth}
    \centering
\includegraphics[height=0.21\textheight]{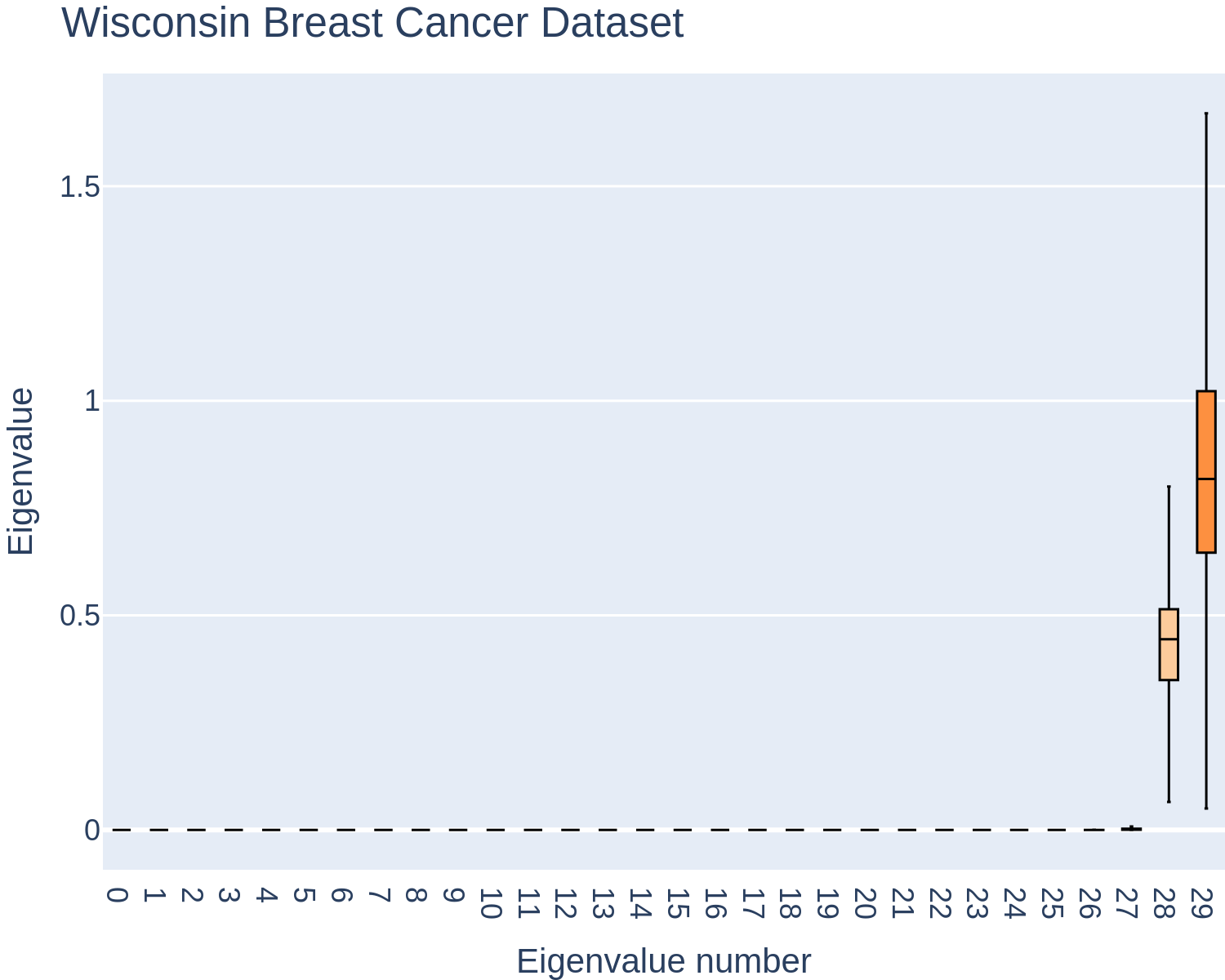}
    \caption{} 
\end{subfigure}
\begin{subfigure}{0.49\textwidth}
    \centering
    \includegraphics[height=0.23\textheight]{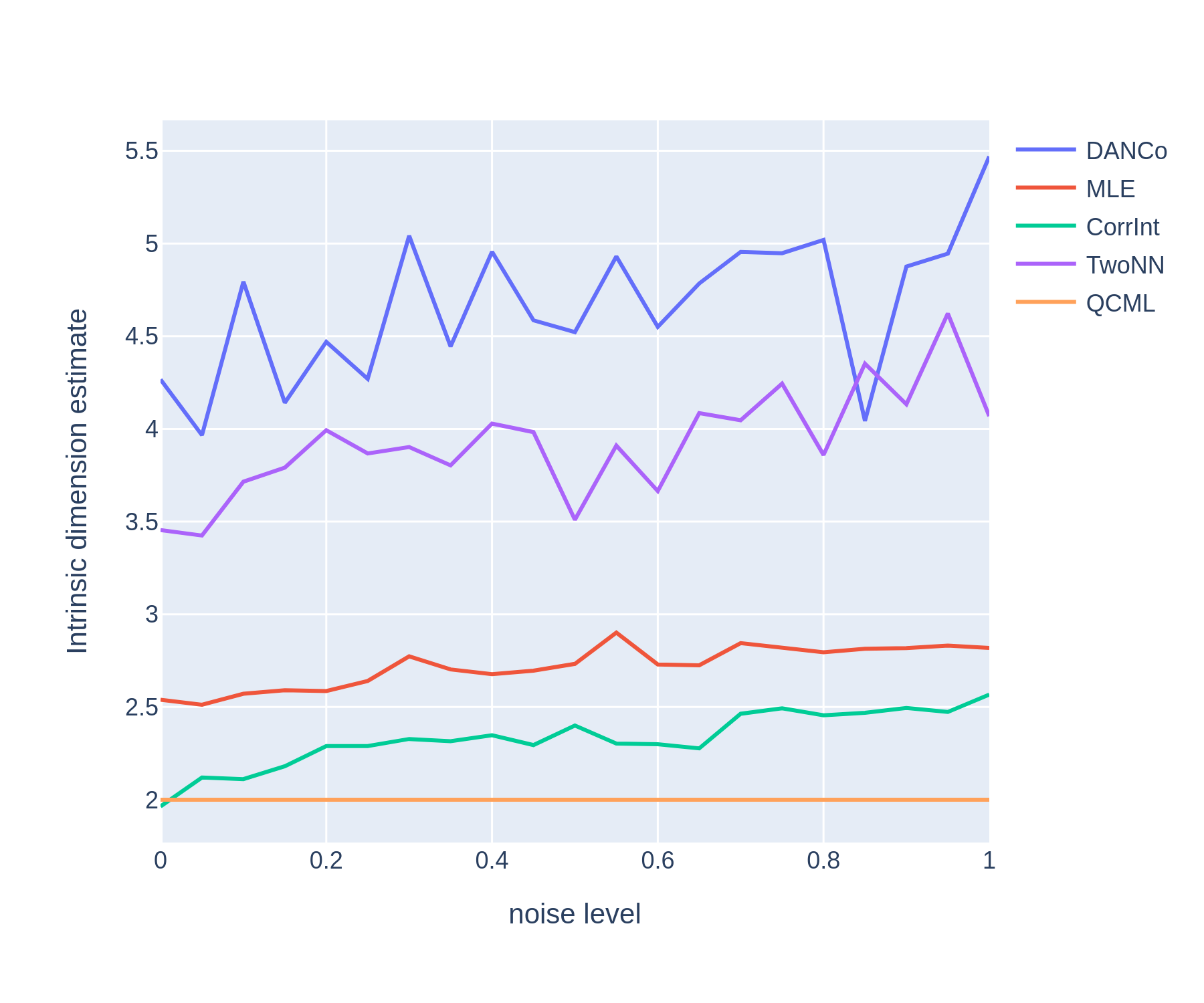}
    \caption{}
\end{subfigure}
\caption{Intrinsic dimension estimates for the Wisconsin Breast Cancer Dataset using a QCML estimator of dimension $N=16$ and quantum fluctuation weight $w=0.1$ in the loss function. (a) Spectral gap with zero \texttt{noise}. Outliers omitted for clarity. (b) Intrinsic dimension estimates as for different estimators as function of \texttt{noise}. For the QCML estimator, a global estimate of dimension is obtained by taking the mode of the local estimates. }
\label{fig:WBC}
\end{figure}

We could not find any previous estimates for the intrinsic dimension of this data set, so we choose $N=16$ for the Hilbert space dimension, according to the rank bound \eqref{eqn:rank_bound}. For the loss function, this time we chose to introduce a quantum fluctuation term with weight factor $w=0.1$, as in \eqref{eqn:matrix_config_optimization_with_variance}. During testing, this choice led to sharper and more consistent spectral gaps (Figure \ref{fig:WBC} (a), showing a spectral gap corresponding to $d=2$). In general, the effects of the quantum fluctuation term on the loss function are analyzed more thoroughly in the appendix at the end of this article.  

To test the robustness of our estimate, we add a synthetic \texttt{noise} parameter consisting of a fraction $\epsilon$ of standard deviation for each feature. That is, if we let $X_1, \ldots, X_{30}$ be the 569-dimensional column vectors representing each feature, we create synthetic noisy features $Y_i$ by letting 
\[
Y_i = X_i + \epsilon\sigma_i \cdot Z_i,
\]
where $\sigma_i$ is the standard deviation for the vector $X_i$ and $Z_i$ is a vector of $N(0,1)$-distributed random entries. In this experiment, the goal is to provide an intrinsic dimension estimate that is constant across different levels of noise $\epsilon$, just like we did for the synthetic manifold examples. 

The results are shown in Figure \ref{fig:WBC} (b), where we tested on 21 equally spaced noise levels from $\epsilon = 0$ to $\epsilon = 1$ in increments of 0.05. 

The QCML estimator consistently returns an intrinsic dimension estimate of $d=2$ across all levels of noise tested. We also plot in Figure \ref{fig:WBC} (b) the results for other estimators. The estimates of these other models tend to slope upwards as the noise level increases, precisely as in the synthetic manifold examples. If we assume that the dataset carries a natural level of noise, then Figure \ref{fig:WBC} suggests that the estimates of all the other methods should be revised downwards, and thus be closer to $d=2$. In fact, not knowing a priori what the level of noise is, we can only imagine extrapolating the graph of Figure \ref{fig:WBC} to the left to a point where all the estimates converge to $d=2$, similar to what happens in synthetic manifolds tests. 

\section*{Discussion}

In this article we introduce a new data representation paradigm based on Quantum Cognition Machine Learning, and  intrinsic dimension estimator  and quantum geometry. The idea is to learn a non-commutative quantum model \cite{Ishiki, Schneiderbauer_2016, Steinacker_2021} for the data manifold itself. This quantum model has the ability to abstract out the fundamental features of the geometry of the data manifold. In particular, we demonstrate how the intrinsic dimension of the data can be estimated from the point cloud produced by the quantum model. Because the point cloud reflects global properties of the data, our method is fundamentally robust to noise, as demonstrated on synthetic benchmarks. This is in contrast to other state-of-the-art techniques, which tend to overestimate intrinsic dimension by including ``shadow'' dimensionality from noise artifacts. In light of our results, we suggest a new paradigm for testing intrinsic dimension estimators: instead of focusing on noise-free synthetic benchmarks of increasing non-linearity and dimensionality, it is perhaps more relevant to focus on the development of estimators that are robust to noise. For practical applications, no real data is immune to noise, and not much meaning can be attached to an intrinsic dimension estimate that is highly dependent on noise levels.

\section*{Methods}

All the results of this article have been obtained by training matrix configurations as in \eqref{eqn:matrix_config_optimization} and \eqref{eqn:matrix_config_optimization_with_variance} on a 32-core 13th Gen Intel Core i9-13950HX CPU with 64GB of memory, supplemented by a NVIDIA RTX 5000 Ada Generation Laptop GPU. Training these models involves iterative updates to the quasi-coherent states $\ket{\psi_0(x)}$ and the matrix configuration $A$ to lower the loss function until desired convergence is obtained. The specifics of each optimization step depend on the particular loss function used and the choice of initialization of the matrix configuration $A$. A typical training loop would consist, for each epoch, of:

\begin{itemize}
\item[(1)] Calculate the quasi-coherent states $\ket{\psi_0(x)}$ for all data points $x \in X$ (or batch of data). 
\item[(2)] Compute the loss function \eqref{eqn:matrix_config_optimization} or \eqref{eqn:matrix_config_optimization_with_variance} and its gradients with respect to $A$. 
\item[(3)] Update the matrix configuration $A$ with gradient descent. 
\end{itemize}

The above training loop was implemented in PyTorch \cite{PyTorch} to obtain all the matrix configurations shown in this article. All other intrinsic dimension estimators (DANCo, MLE, CorrInt, TwoNN) were tested through their implementation in the \texttt{scikit-dimension} Python package \cite{scikit-dimension}.

\bibliographystyle{alpha}
\bibliography{intrinsic_dimension.bib}

\section*{Appendix}

\subsection*{Effect of quantum fluctuation control on loss function}

\begin{figure}[H]
\centering

\begin{subfigure}{0.49\textwidth}
    \centering
    \includegraphics[scale=0.5]{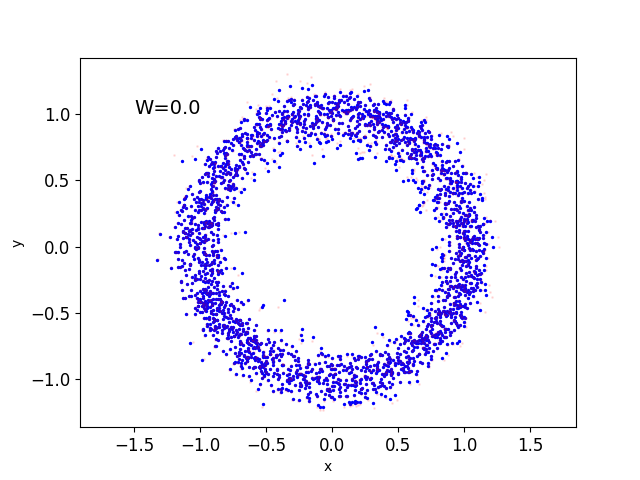}
    \caption{} 
\end{subfigure}
\begin{subfigure}{0.49\textwidth}
    \centering
    \includegraphics[scale=0.5]{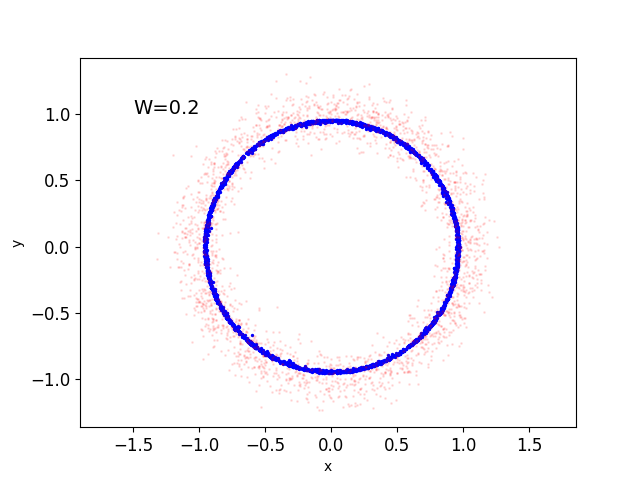}
    \caption{}

\medskip
    
\end{subfigure}
\begin{subfigure}{0.49\textwidth}
    \centering
    \includegraphics[scale=0.5]{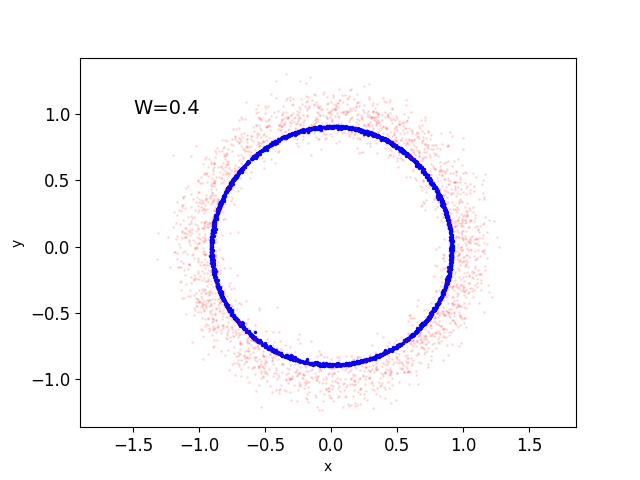}
    \caption{}
\end{subfigure}
\begin{subfigure}{0.49\textwidth}
    \centering
    \includegraphics[scale=0.5]{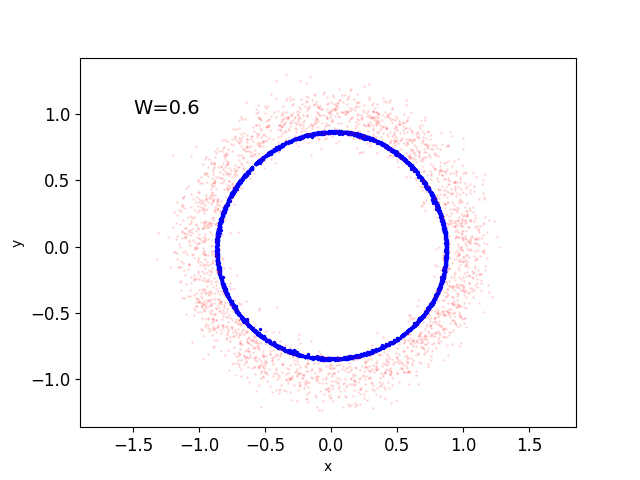}
    \caption{} 
\end{subfigure}

\medskip

\begin{subfigure}{0.49\textwidth}
    \centering
    \includegraphics[scale=0.5]{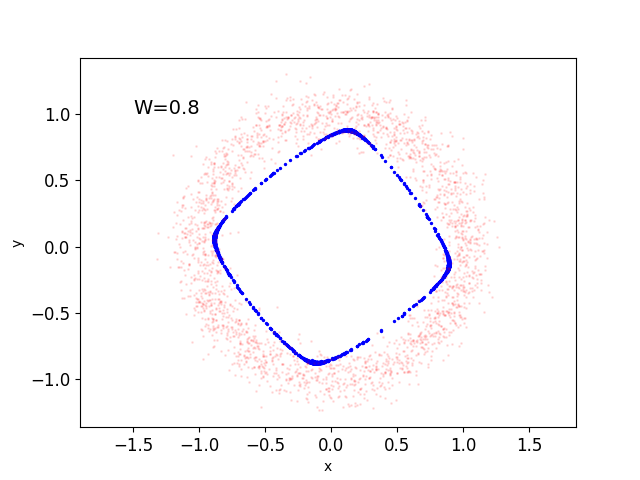}
    \caption{}
\end{subfigure}
\begin{subfigure}{0.49\textwidth}
    \centering
    \includegraphics[scale=0.5]{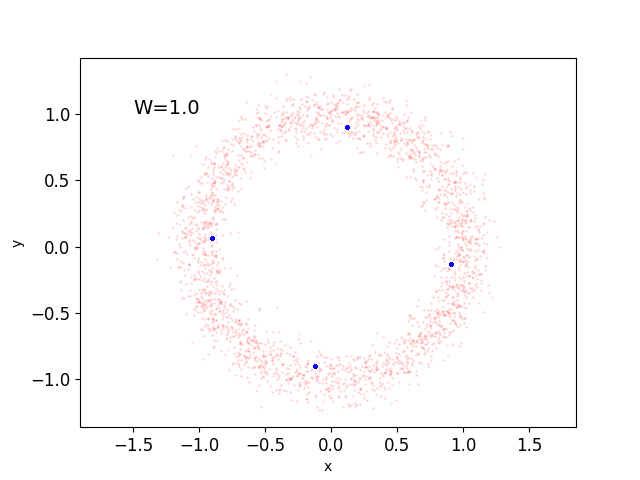}
    \caption{}
\end{subfigure}

\caption{Dataset consisting of $T=2500$ uniformly distributed points on a unit circle with \texttt{noise}=0.1, shown in red. Six different point clouds are shown (in blue) corresponding to six different levels of quantum fluctuation weight $w$ in the loss function \eqref{eqn:matrix_config_optimization_with_variance}. The Hilbert space dimension is $N=4$.}
\label{fig:unit_circle}
\end{figure}

Consider a matrix configuration $A$ trained on the data set $X$ according to the loss function \eqref{eqn:matrix_config_optimization_with_variance}
\[
A = \displaystyle\mathrm{argmin}_{B=\{B_1, \ldots, B_D\}} \left( \sum_{x\in X} {\lVert B(\psi_0(x)) - x\rVert^2} + w\cdot \sigma^2(x) \right),
\]
where $w$ is a real parameter $0 \leq w \leq 1$ controlling the quantum fluctuation term. In Figure \ref{fig:unit_circle} we show the effect on the point cloud $X_A$ of varying $w$, for the case when $X$ is a synthetic dataset consisting of 2500 points uniformly distributed on the unit circle, with a Gaussian $\texttt{noise}$ parameter of 0.1. For this example, we used a QCML estimator of Hilbert space dimension $N=4$. For $w=0$ the point cloud (shown in blue) is an almost perfect match to the original data, including the noise. The manifold approximation in this case is low bias/high variance. As $w$ ranges from 0.2-0.6, the point cloud instead resembles closely the unit circle, filtering out most of the noise. The bias is higher compared to the case $w=0$, but the variance is lower. As $w$ increases, the point cloud begins however to degenerate into a 4-means clustering approximation of the data manifold: in this case, this just corresponds to the vertices of square inscribed inside the unit circle. This example shows how the quantum fluctuation term in \eqref{eqn:matrix_config_optimization_with_variance} controls the variance of the point cloud approximation $X_A$: while low levels of quantum fluctuation control lead the model to filter out noise and highlight important features of the data, training with the full energy loss function (i.e $w=1$) may lead to undesirable degenerations of the matrix configuration $A$.

\newpage

\subsection*{Controlling variance by varying the Hilbert space dimension}

\begin{figure}[H]
\centering

\begin{subfigure}{0.49\textwidth}
    \centering
    \includegraphics[scale=0.11]{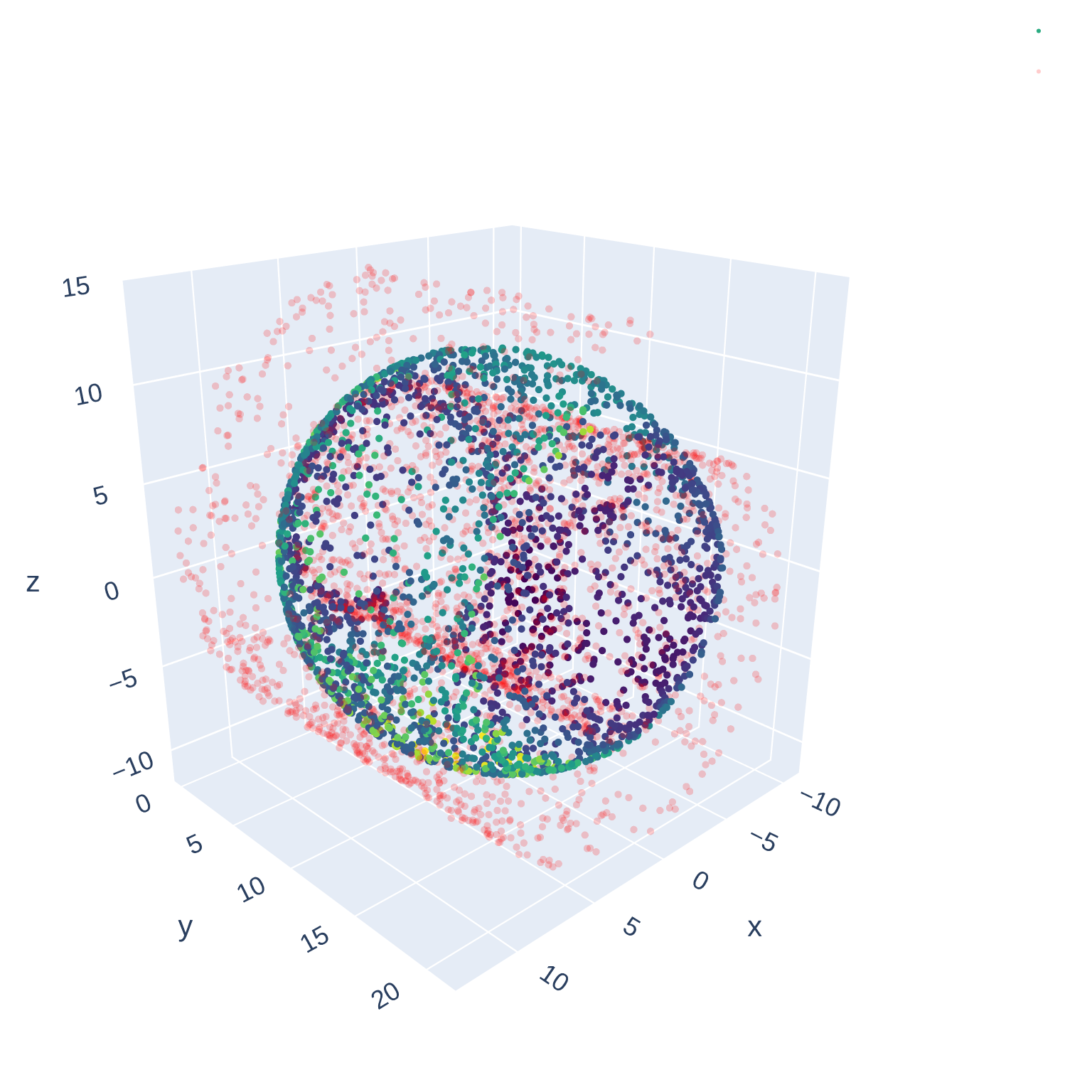}
    \caption{} 
\end{subfigure}
\begin{subfigure}{0.49\textwidth}
    \centering
    \includegraphics[scale=0.11]{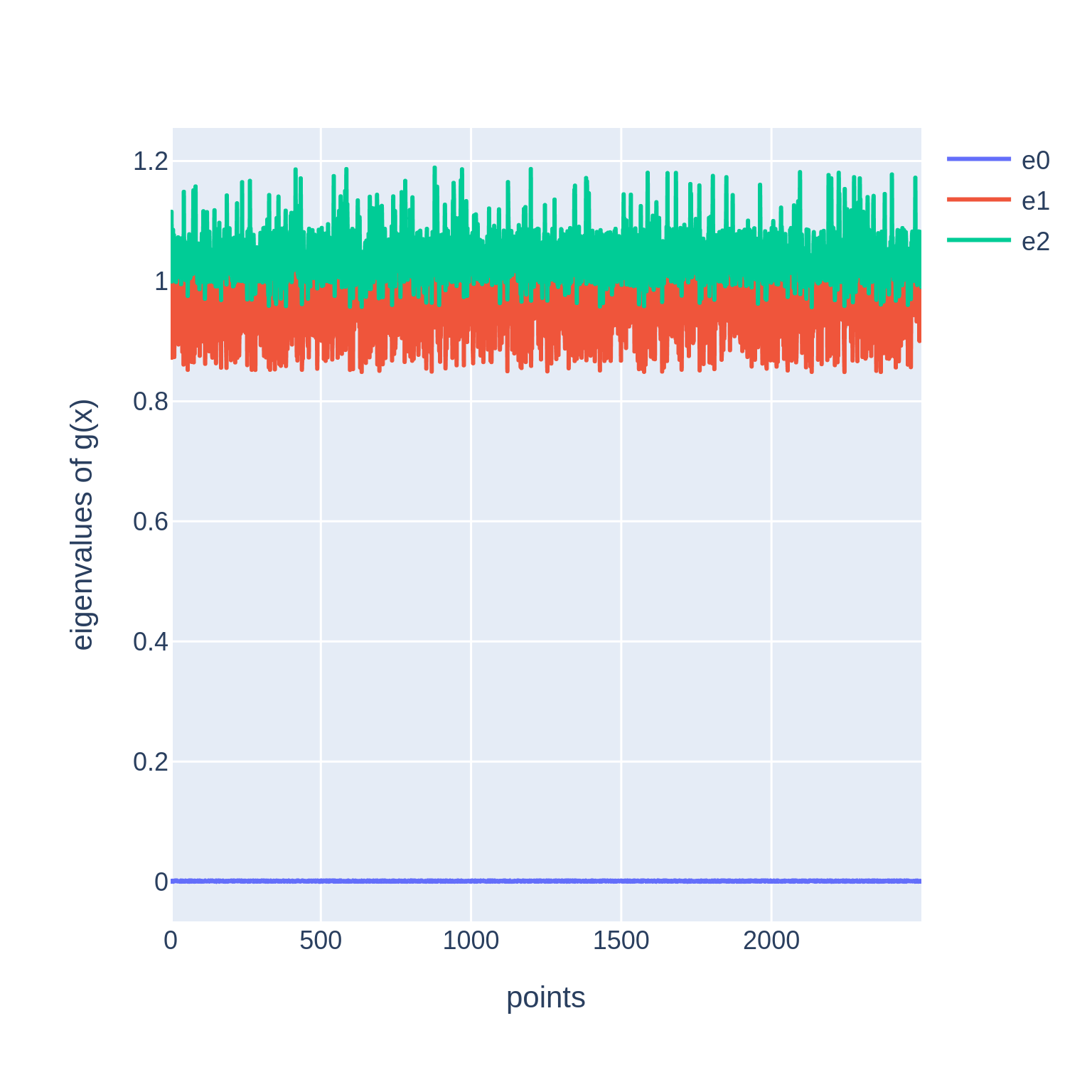}
    \caption{}
\end{subfigure}

\medskip

\begin{subfigure}{0.49\textwidth}
    \centering
    \includegraphics[scale=0.11]{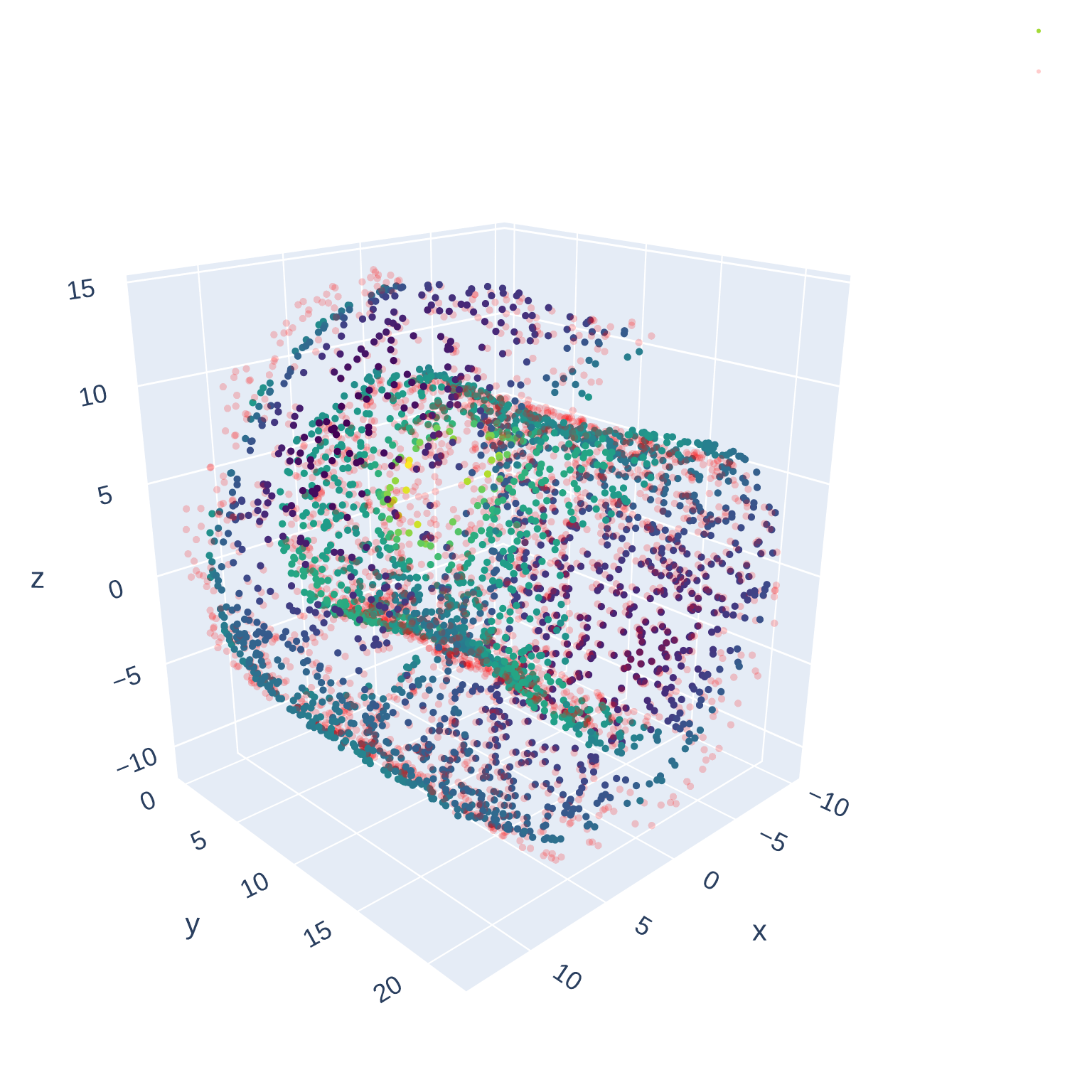}
    \caption{} 
\end{subfigure}
\begin{subfigure}{0.49\textwidth}
    \centering
    \includegraphics[scale=0.11]{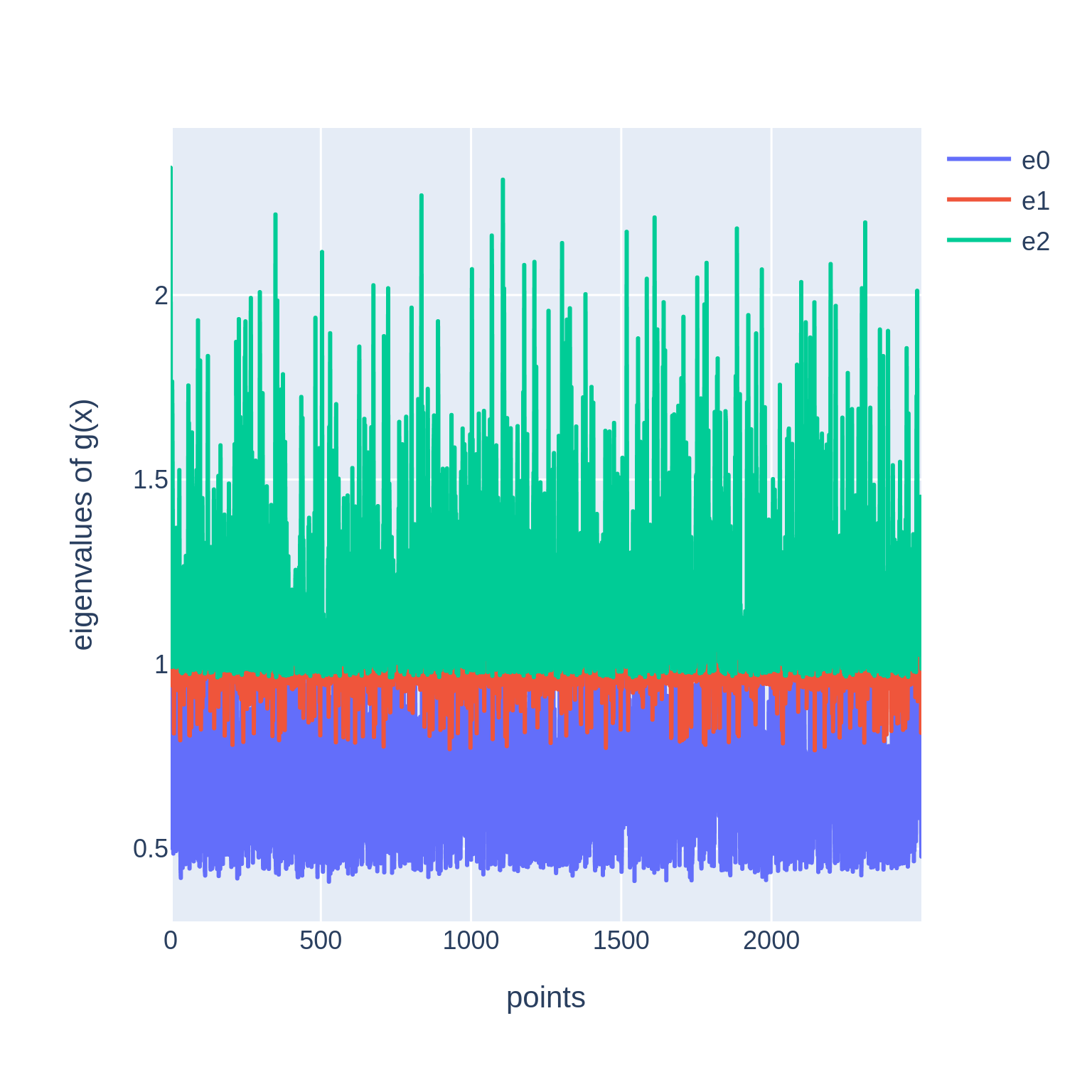}
    \caption{}
\end{subfigure}

\caption{Dataset consisting of $T=2500$ points on a Swiss roll with zero noise. Two configurations are shown, (a-b) point cloud approximation and spectral gap for Hilbert space dimension $N=3$ and (c-d) for $N=4$. The original data set is in red. Darker colors in the point cloud correspond to lower energy.  }
\label{fig:swiss_roll}
\end{figure}

Another way to control variance of the point cloud approximation is to vary the Hilbert space dimension $N$. In Figure \ref{fig:swiss_roll} we show the results of training a QCML estimator of dimension $N=3$ and $N=4$ on a `Swiss roll' synthetic data set. This manifold is labeled $M7$ in the \texttt{scikit-dimension} Python package \cite{scikit-dimension} that we use for benchmarking. In these examples, we kept $w=0$ and only trained the bias term of the loss function, with no quantum fluctuation control. For $N=3$ the point cloud approximation looses quite a bit of information about the global geometry of the manifold, but the spectral gap between the eigenvalues of the quantum metric is very well-defined, and it returns the correct intrinsic dimension estimate $d=2$ at every point (Figure \ref{fig:swiss_roll} (a-b)).  Increasing the Hilbert space dimension to $N=4$ allows the point cloud to gain expressivity and better match the original data set (Figure \ref{fig:swiss_roll} (c)) but a less clear spectral gap estimate is obtained. In this example, 807 points returned an intrinsic dimension estimate of $d=1$ and 1693 points returned $d=2$. This ambiguity is reflected in the graph of Figure \ref{fig:swiss_roll} (d)). 

\end{document}